\documentclass[10pt,twocolumn,letterpaper]{article}

\usepackage[pagenumbers]{cvpr} %

\usepackage[dvipsnames,table]{xcolor}

\usepackage{siunitx}  %
\usepackage{bbm}
\usepackage{multirow}

\usepackage{booktabs}
\usepackage[normalem]{ulem}
\useunder{\uline}{\ul}{}

\definecolor{ms_note}{RGB}{0, 181, 190}
\definecolor{bgcolor}{RGB}{190, 181, 190}
\definecolor{purple}{RGB}{255, 20, 140}
\newcommand{\closecompetitor}[1]{#1}

\usepackage{pifont}

\definecolor{mylightgray}{RGB}{238,238,238} 
\colorlet{bgcolor}{mylightgray}

\newcommand{\opacity}{\sigma}
\newcommand{\mean}{\mu}

\newcommand{\colorsh}{c} %

\newcommand{\code}{\mathbf{h}}  %

\renewcommand{\paragraph}[1]{\vspace{0pt}\noindent\textbf{#1}\hspace{6pt}}

\definecolor{cvprblue}{rgb}{0.21,0.49,0.74}
\usepackage[pagebackref,breaklinks,colorlinks,citecolor=cvprblue]{hyperref}

\def\paperID{226} %
\def\confName{3DV\xspace}
\def\confYear{2026\xspace}

\title{Complete Gaussian Splats from a Single Image with Denoising Diffusion Models}

\author{Ziwei Liao\textsuperscript{1} \quad Mohamed Sayed\textsuperscript{2} \quad Steven L. Waslander\textsuperscript{1} \\ 
Sara Vicente\textsuperscript{2} \quad Daniyar Turmukhambetov\textsuperscript{2} \quad Michael Firman\textsuperscript{2}\\
\textsuperscript{1}University of Toronto \quad \textsuperscript{2}Niantic Spatial\\
\\
\href{https://nianticspatial.github.io/completesplat}{https://nianticspatial.github.io/completesplat}
}

\begin{document}
\maketitle

\begin{abstract}

Gaussian splatting typically requires dense observations of the scene and can fail to reconstruct occluded and unobserved areas. 
We propose a latent diffusion model to reconstruct a \textit{complete} 3D scene with Gaussian splats, including the occluded parts, from only a single image during inference.
Completing the unobserved surfaces of a scene is challenging due to the ambiguity of the plausible surfaces.  
Conventional methods use a regression-based formulation to predict a single ``mode'' for occluded and out-of-frustum surfaces, leading to blurriness, implausibility, and failure to capture multiple possible explanations.  
Thus, they often address this problem partially, focusing either on objects isolated from the background, reconstructing only visible surfaces, or failing to extrapolate far from the input views.  
In contrast, we propose a \textit{generative} formulation to learn a distribution of 3D representations of Gaussian splats conditioned on a single input image.  
To address the lack of ground-truth training data, we propose a Variational AutoReconstructor to learn a latent space only from 2D images in a self-supervised manner, over which a diffusion model is trained.
Our method generates faithful reconstructions and diverse samples with the ability to complete the occluded surfaces for high-quality 360$^\circ$ renderings.

\end{abstract}

\section{Introduction}

\begin{figure}
    \centering
    \includegraphics[width=1.0\linewidth]{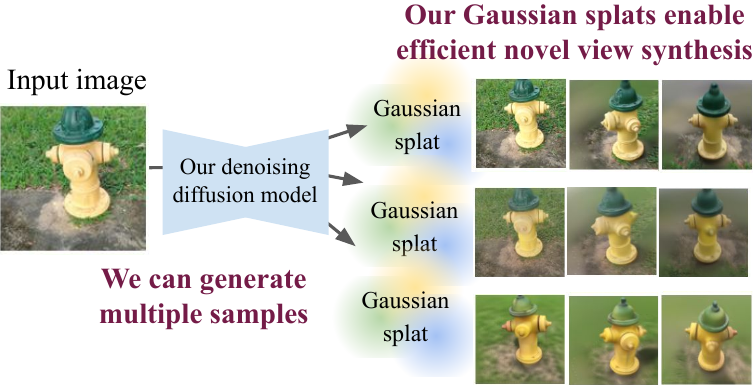}
    \caption[Full Gaussian scenes prediction from a single RGB input image.]{
        \textbf{We predict full Gaussian scenes from a single RGB input image.}
        Our diffusion-based model outputs sharper results than existing methods, and is also able to sample diverse completion ``modes'' given a single image as input.
    }
    \vspace{-10pt}
    \label{fig:gaussian_title_figure}
\end{figure}

Gaussian splatting~\cite{kerbl3Dgaussians} has demonstrated impressive performance in modeling 3D scenes, delivering high-quality renderings with applications in augmented reality, world simulations, and robotics.
Traditionally, dense observations, typically hundreds of RGB images, are required to accurately capture scene details and constrain the reconstruction process~\cite{mildenhall2020nerf,kerbl3Dgaussians}, due to the large number of unknown parameters.
However, such dense observations are not always available, often resulting in blurriness, artifacts, or empty regions caused by underconstrained reconstructions in occluded or invisible areas, such as the backs of objects and regions outside the camera frustum.
In cases where sparse or even a single view is available, prior knowledge beyond the observations becomes critical for constraining the reconstruction.

We aim to address the learning of prior knowledge from large-scale datasets to help constrain the highly ill-posed problem of reconstructing a complete 3D scene from a single RGB image. 
Researchers have trained neural networks to reconstruct 3D Gaussian splats from a single image~\cite{szymanowicz2024splatter,szymanowicz2024flash3d} or sparse two-view images~\cite{charatan2024pixelsplat,hen2024mvsplat}. 
These approaches can be broadly categorized into two formulations: \textit{regression-based} and \textit{generative}.  
As \textit{regression-based} methods, Splatter Image~\cite{szymanowicz2024splatter} and its follow-up works~\cite{szymanowicz2024flash3d,charatan2024pixelsplat,hen2024mvsplat} predict multiple Gaussians along each camera ray in a pixel grid using a single forward pass, efficiently modeling scenes within the image frustum. 
However, a fundamental limitation exists for regression-based formulations. They produce \textit{unimodal} predictions, where only a single output is predicted for a given input image.
This forces the model to average multiple hypotheses into a single ``mode'' for occluded or ambiguous regions, often resulting in blurry or inaccurate reconstructions.
Due to this limitation, Splatter Image demonstrates performance only on isolated foreground objects. 
While follow-up works~\cite{szymanowicz2024flash3d,charatan2024pixelsplat,hen2024mvsplat} extend these methods to scene-level reconstruction, they remain constrained to simpler tasks such as interpolation and exhibit limited extrapolation ability, particularly when generating parts of the scene that are not directly visible.

Instead, we propose a \textit{generative} approach to learn a distribution over 3D Gaussian splats, enabling the sampling of multiple high-quality hypotheses in ambiguous situations, as shown in Figure~\ref{fig:gaussian_title_figure}.
Generative formulations have also been explored on NeRFs~\cite{tewari2023diffusion} and Gaussian splats~\cite{wewer24latentsplat,henderson2024sampling}. 
Different from LatentSplat~\cite{wewer24latentsplat}, which follows a variational autoencoder (VAE), we leverage more powerful denoising diffusion models~\cite{ho2020denoising}, which have demonstrated a stronger ability to model complex distributions in images~\cite{rombach2022stablediffusion} and 3D models~\cite{jun2023shap}, to better capture the complex distributions of Gaussian splatting.

Recently, diffusion models have been used for Gaussian splatting. 
Some approaches~\cite{chen2024text,yi2024gaussiandreamer,tang2023dreamgaussian,li2023gaussiandiffusion} leverage priors from 2D image diffusion models to generate 3D Gaussians, which requires intensive optimization and can suffer from consistency issues. 
Limited work exists that learns diffusion models to directly model \textit{3D distributions} of splats. 
For example, Bolt3D~\cite{szymanowicz2025bolt3d} relies on extra regression heads for generating Gaussians from denoised intermediate outputs.
Some methods require ground truth 3D data of Gaussian splats~\cite{xiang2025repurposing, meng2025zero}, which is expensive to obtain and limits scalability to diverse scenes.
DFM~\cite{tewari2023diffusion} trains a denoising diffusion model over NeRFs~\cite{mildenhall2020nerf} from only image supervision with forward models. 
Its architecture requires computationally expensive denoising for each rendered image and also inherits NeRF's high computational cost.  

Different from the above, we propose a novel generative method for 3D Gaussian splatting based on a latent diffusion architecture~\cite{rombach2022stablediffusion} to directly learn the distribution of 3D Gaussian splats in a latent space from \textit{only image supervision}, enabling both real-time computation and high-quality, diverse samplings.
This is not a straightforward process, and we make the following critical contributions to enable an effective training pipeline.
The first challenge is how to acquire 3D ground truth Gaussian splat training data. To address this, we propose a novel method, Variational AutoReconstructor, to learn a latent space for 3D Gaussian splats using self-supervised losses derived solely from images.
To mitigate the loss of high-frequency details in the latent space, we use skip connections to propagate and preserve fine-grained details.
We further propose an approach based on classifier-free guidance to provide flexible control that balances faithfulness and randomness in the outputs for different applications.
Our contributions are as follows:

\begin{itemize}
    \item We propose a generative Gaussian splatting reconstruction method with a latent diffusion model for generating complete 3D scenes from a single RGB image, enabling the completion of occluded surfaces with multi-view consistency and efficient reasoning.
    \item We present a novel Variational AutoReconstructor to learn the latent space for Gaussian splats from only 2D posed images using self-supervised differentiable rendering losses.
    \item We present flexible control over the fidelity of reconstructed visible surfaces and the diversity of generated occluded surfaces within a latent diffusion model, using skip connections and classifier-free guidance.
\end{itemize}

Our experiments on the Hydrants and TeddyBears categories from the CO3D~\cite{reizenstein21co3d} dataset, as well as room scenarios from the RealEstate10K~\cite{zhou2018stereo} dataset, show that our predictions are sharper and more complete than those of state-of-the-art methods, and provide new abilities to sample diverse 3D outputs.

\section{Related Work}

There is limited work on training denoising diffusion models to directly output 3D Gaussian splats.  
We provide an overview highlighting the differences with the broader topic in Table~\ref{tab:related_work}, and discuss them in detail below.

\paragraph{Summary}
Unlike novel view synthesis methods that produce only 2D images~\cite{gu2023nerfdiff,chan2023genvs,hollein2024viewdiff}, or videos lacking explicit 3D modeling and consistency~\cite{ho2022imagen,he2025cameractrl}, or virtual scenes generated by text-to-3D models that suffer from domain gaps compared to real scenes~\cite{yi2024gaussiandreamer,tang2023dreamgaussian}, our model naturally learns and outputs distributions of 3D Gaussian splats of real scenes with guaranteed 3D geometric consistency.  
Unlike object-only reconstruction models~\cite{xu2024agg,zhang2024geolrm,boss2024sf3d,zhou2023sparsefusion}, we reconstruct complete 3D scenes, including both foreground objects and background.  
Different from offline NeRF-based methods~\cite{tewari2023diffusion}, our representation can be rendered in real-time.  
And unlike regression-based methods~\cite{szymanowicz2024splatter,charatan2024pixelsplat,hen2024mvsplat}, we use a \textit{generative} formulation with a denoising diffusion model, enabling the completion of occluded or invisible parts and the sampling of diverse scene variations.

\begin{table*}
    \centering
    \newcommand{\yes}{\checkmark}
    \resizebox{1.0\textwidth}{!}{
\begin{tabular}{@{}llllllll@{}}
\toprule
\textbf{Scale}           & \textbf{Methods}             & \textbf{Outputs}                             & \textbf{Inputs}                & \textbf{Formulations}                                & \textbf{Architectures}            & \textbf{3D Representations} & \textbf{Computation}              \\ \midrule
Objects                  & Objects Models~\cite{xu2024agg,zhang2024geolrm,boss2024sf3d,zhou2023sparsefusion}               & 3D                                           & /                              & /                                                    & /                                 & GS, NeRF                    & /                                 \\ \midrule
                         & Novel View Sythesis~\cite{gu2023nerfdiff,chan2023genvs,hollein2024viewdiff}          & Image                                        & /                              & /                                                    & /                                 &                             & /                                 \\
                         & Video Generation~\cite{ma2025you,liang2025wonderland}             & Video                                        & /                              & /                                                    & /                                 & \multirow{-2}{*}{None}      & /                                 \\ \cmidrule(l){2-8} 
                         & SplatterImage~\cite{szymanowicz2024splatter,szymanowicz2024flash3d,charatan2024pixelsplat,hen2024mvsplat}          & 3D                         & 1 or 2-views                        & Regression-based                   & /                                 & GS        & Real-time       \\ \cmidrule(l){2-8} 
                         & LatentSplat~\cite{wewer24latentsplat}                  & \cellcolor[HTML]{EFEFEF}                     & 2-views                        & \cellcolor[HTML]{EFEFEF}                             & VAE                               & GS                          & Real-time                         \\
                         & CAT3D~\cite{gao2024cat3d}                   & \cellcolor[HTML]{EFEFEF}                     & \cellcolor[HTML]{EFEFEF}1-view                        & \cellcolor[HTML]{EFEFEF}                             & 2D Diffusion                               & GS                          & Offline                         \\
                         & DFM~\cite{tewari2023diffusion}                          & \cellcolor[HTML]{EFEFEF}                     & \cellcolor[HTML]{EFEFEF}1-view & \cellcolor[HTML]{EFEFEF}                             & \cellcolor[HTML]{EFEFEF}3D Diffusion & NeRF                        & Offline                           \\
\multirow{-7}{*}{Scenes} & \cellcolor[HTML]{EFEFEF}Ours & \multirow{-4}{*}{\cellcolor[HTML]{EFEFEF}3D} & \cellcolor[HTML]{EFEFEF}1-view & \multirow{-4}{*}{\cellcolor[HTML]{EFEFEF}Generative} & \cellcolor[HTML]{EFEFEF}3D Diffusion & \cellcolor[HTML]{EFEFEF}GS  & \cellcolor[HTML]{EFEFEF}Real-time \\ \bottomrule
\end{tabular}
    }
    \caption[Comparison to Related Works.]{\textbf{Comparison to Related Works.} The table highlights the main differences from closely related baselines. We propose a generative method with diffusion models to reconstruct 3D scenes with Gaussian splats in real time from a single image.}
    \label{tab:related_work}
    \vspace{-15pt}
\end{table*}

\subsection{Learning 3D Reconstruction and Generation}

\noindent
\textbf{Object-centric Models}~\cite{xu2024agg,zhang2024geolrm,boss2024sf3d,zhou2023sparsefusion,xu2023dmv3d,liu2024novelgs,xu2024grm,shen2025gamba} reconstruct and generate 3D \textit{object} models given text prompts or images, but lack awareness of background and scene context.  
Due to the domain gap in large-scale object datasets~\cite{deitke2023objaverse}, the generated outputs often appear synthetic and virtual, with noticeable discrepancies from real-world textures and geometry.  
Notably, DMV3D~\cite{xu2023dmv3d} denoises a triplane NeRF representation, whereas ours denoises Gaussian splatting in a latent space, enabling much faster inference and training. 
Most importantly, it remains unclear how well these methods scale to larger and more complex scene-level reconstruction tasks, especially in the absence of large-scale, ground-truth 3D scene data.

\noindent
\textbf{Scene-level ``Regression" Models} predict a single ``mode'' of the 3D representation from partial input data. However, in ambiguous situations, such as occluded areas, the network tends to learn an ``average'' mode, which often leads to blurry results.
Early works focused on voxel grids~\cite{wu20153d,song2017semantic,wang2017shape,dai2018scancomplete,dai2020sg,dai2017shape,galvis2024sc} and occupancy fields~\cite{huang2024zeroshape}, which model geometry only, without capturing texture. 
Dust3R and its variants~\cite{wang2024dust3r,wang2025vggt} show promising end-to-end learning for camera localization and reconstruction with point clouds, which is orthogonal to ours, where we focus on representing 3D scenes.
To synthesize 360$^\circ$ views with NeRFs~\cite{mildenhall2020nerf}, PixelNeRF~\cite{yu2021pixelnerf} predicts novel views conditioned on features extracted from single (or few) input images. NViST~\cite{jang2024nvist} further extends this with a Transformer to predict novel views.  
However, these methods are limited by the slow rendering process of NeRF, which is mitigated by Gaussian splats~\cite{kerbl3Dgaussians}.
Splatter Image~\cite{szymanowicz2024splatter} and follow-up works~\cite{fei2024pixelgaussian,hong2024pf3plat} predict multiple Gaussians along camera rays in a pixel grid, resulting in efficient methods to model scenes within image frustums. Multi-view inputs~\cite{charatan2024pixelsplat}, depth priors~\cite{szymanowicz2024flash3d,liu2025monosplat,xu2024depthsplat}, and structural constraints~\cite{hen2024mvsplat} have been further explored to regularize the outputs.

\noindent
\textbf{Scene-level ``Generative" Models} can predict a ``distribution'' of 3D scene representations to model complex and ambiguous situations.  
LatentSplat~\cite{wewer24latentsplat} follows a variational autoencoder (VAE) design to infer the distribution of Gaussian splats.  
We leverage more powerful denoising diffusion models~\cite{ho2020denoising}, which have demonstrated a stronger ability to model complex distributions in images~\cite{rombach2022stablediffusion} and 3D models~\cite{jun2023shap}, to better capture the complex distributions of Gaussians. 
DFM~\cite{tewari2023diffusion} trains a denoising diffusion model to learn a distribution of NeRFs conditioned on a single input image. However, it requires running denoising steps for \textit{each} novel view, requiring extremely intensive computation.

\noindent
\textbf{Novel View Synthesis and Video Generation Models} output \textit{images} or \textit{videos}~\cite{ma2025you,liang2025wonderland} conditioned on camera poses.  
NerfDiff~\cite{gu2023nerfdiff}, GeNVS~\cite{chan2023genvs}, and ViewDiff~\cite{hollein2024viewdiff} train diffusion models to generate novel views.  
However, they cannot directly output a coherent 3D representation. 
Instead, they need to run the intensive denoising process separately for novel views to generate images or videos, which increases computational cost and, most importantly, limits multi-view consistency without 3D geometry.

\subsection{Diffusion Models for Gaussian Splats}

\paragraph{2D Diffusion: Lifting Image Diffusion Models for 3D}  
2D image generation models~\cite{rombach2022stablediffusion} contain priors that can be used to optimize 3D representations using Score Distillation Sampling (SDS) from rendered multi-view images of NeRFs~\cite{poole2022dreamfusion,sargent2024zeronvs} or Gaussian splats~\cite{chen2024text,yi2024gaussiandreamer,tang2023dreamgaussian,li2023gaussiandiffusion}.  
Some approaches first synthesize multiple novel view images and then optimize the 3D scene~\cite{feng2024fdgaussian,liu2023zero,gao2024cat3d}.  
However, a large number of novel views are required to fully constrain 3D representations during optimization, where identity and multi-view consistency are challenging, requiring finetuning of the large diffusion model (\eg Zero 1-to-3~\cite{liu2023zero} and Cat3D~\cite{gao2024cat3d}). 
Some methods~\cite{zhang2025scene,schwarz2025generative} use video diffusion models as priors.
In principle, the priors of image and video generation models operate in 2D image space rather than directly on the 3D representation. As a result, these methods suffer from long optimization times and geometric consistency issues, such as the multi-face problem~\cite{poole2022dreamfusion}.

\paragraph{3D Diffusion: Direct Modelling of 3D Gaussian Splats Distributions} 
There is limited work exploring diffusion models to directly learn the distribution of 3D Gaussian splats, with the biggest challenge being the lack of large-scale 3D ground-truth Gaussian splat datasets for real scenes.  
Some works focus on objects~\cite{meng2025zero,jun2023shap,mu2024gsd,liao2024toward} rather than entire scenes.  
Some works finetune diffusion models originally designed for image generation~\cite{xiang2025repurposing,lin2025diffsplat,meng2025zero}, tailoring the output to align with Gaussian splat modalities. 
However,~\cite{xiang2025repurposing,lin2025diffsplat,meng2025zero} require ground truth 3D Gaussian splats in order to learn the latent space, making these methods computationally intensive due to the need to build a dataset in advance. 
In contrast, we learn the latent space directly from images without requiring ground truth splats, with the potential to scale up with the availability of large-scale image datasets.
Bolt3D \cite{szymanowicz2025bolt3d} uses a multiview diffusion model to generate images and requires an additional intermediate model to produce a splat representation.
Some works~\cite{peng2025lesson} rely on an extra regression-based model as a teacher to supervise the diffusion model, complicating the process.
Most similar to ours, \closecompetitor{Henderson \etal}~\cite{henderson2024sampling} train a denoising diffusion model to generate Gaussian splats directly.
Their model is trained to encode \textit{multi-view} input images into a multi-view latent space. 
However, during inference, it requires knowledge of multiple camera positions, which limits its applicability in real-world scenarios, whereas ours only requires single-view input.

\section{Method}

In this section, we first discuss the challenges of training denoising diffusion models on 3D Gaussian splats \textit{in the absence of ground-truth 3D data}.  
We then propose a novel architecture, the Variational AutoReconstructor, which learns a latent space using forward models (differentiable renderings), along with critical modifications for preserving high-frequency details. 
These components together enable the training of a diffusion model to generate 3D Gaussian splats.

\subsection{Training Diffusion Models without Ground Truth Samples}

Conventional denoising diffusion models~\cite{ho2020denoising,rombach2022stablediffusion} require ground-truth samples to which noise is progressively added, and from which the denoiser learns to remove noise.  
We follow Splatter Image~\cite{szymanowicz2024splatter} to model 3D Gaussian splats within an image frustum by predicting multiple Gaussians along each camera ray.  
In order to train a diffusion model to learn a distribution \( P(X) \) over the Splatter Image \( X \), we need access to a large-scale dataset of \( X \), i.e., ground-truth Splatter Images.  
However, unlike images~\cite{rombach2022stablediffusion}, videos~\cite{ho2022imagen}, or 3D objects~\cite{jun2023shap}, no large-scale scene-level datasets exist for 3D Splatter Image representations, which are costly and time-consuming to construct.
One approach might be to pre-optimize Splatter Images from multi-view inputs~\cite{xiang2025repurposing}. However, aside from the computational and scalability limitations, our early experiments with this method revealed sparsity and continuity issues hinder the effective learning of neural networks.  
Therefore, we propose a more scalable method that directly learns from image datasets~\cite{reizenstein21co3d,dai2017scannet,zhou2018stereo}, without relying on intermediate modules.

\paragraph{Training Diffusion Models over Gaussian splats from \textit{Images Supervision}}
The feasibility of training diffusion models from low-dimensional observations has been scarcely explored in the literature. 
Theoretically, our goal is to learn a diffusion model that approximates the distribution \( P(X) \) without access to ground truth samples of \( X \), but only to indirect projections \( f(X) \) of \( X \). Here, \( f \) is a differentiable function—referred to as a ``forward model'' or ``observation model'', which outputs a lower-dimensional partial projection of the high-dimensional variable \( X \).
In our case, the projections are rendered images from Gaussian splats. This links our image datasets of scenes to the potential unknown Gaussian representations that can model those scenes. 
Instead of directly training on Splatter Images, we follow a latent diffusion architecture, which models the distribution over the latent space.  
The latent space learns and compresses the 3D structure of Splatter Images and reduces their high-dimensional parameters, enabling much more computationally efficient training of the diffusion model.

\subsection{Learning a Latent Space of 3D Splatter-Images from 2D Images}
\label{sec:training_step1}

\begin{figure}
    \centering
    \includegraphics[width=0.42\textwidth]{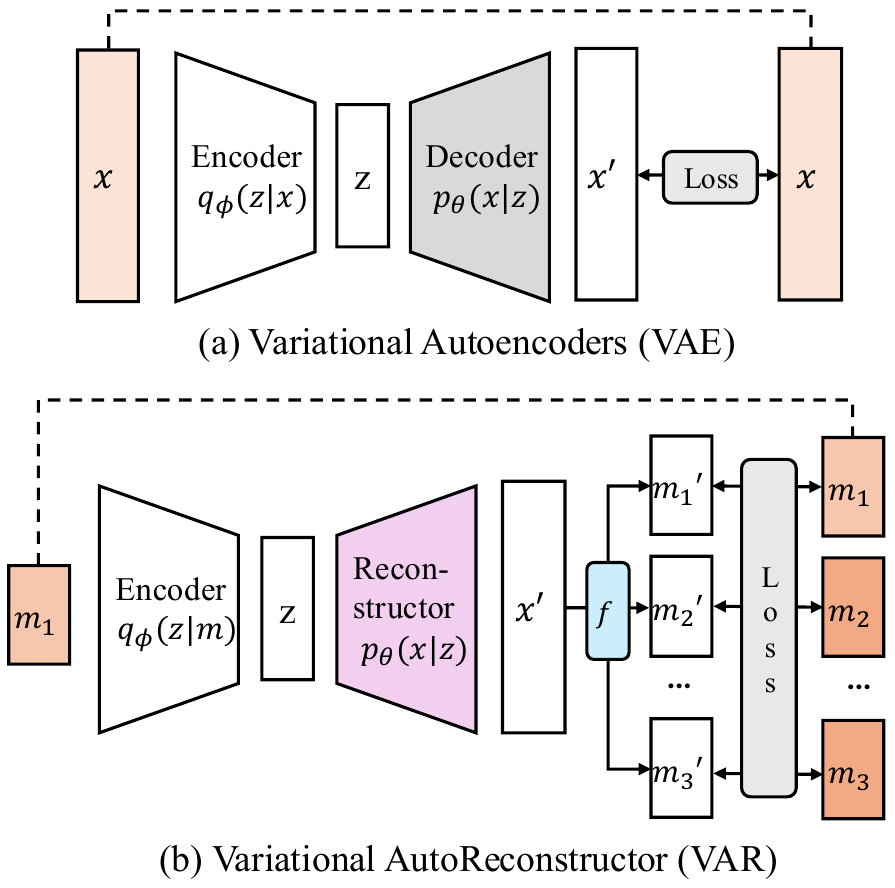}
    \caption[Variational Autoencoders and Variational AutoReconstructor.]{
    \textbf{Learning a latent space for 3D representations using only images, without ground-truth 3D data.}
    (a) Variational Autoencoders require groundtruth samples of high-dimensional variables \( x \) to learn a latent space;  
    (b) We propose the \textit{Variational AutoReconstructor}, which learns a latent space for \( x \) using supervision from only their projections \( \{m = f(x)\} \).
    }
    \vspace{-10pt}
    \label{fig:autoreconstructor}
\end{figure}

We propose a novel architecture, \textit{Variational AutoReconstructor}, to learn the latent space in a self-supervised manner from only images, over which a denoising diffusion model is then trained.
We discuss alternative methods that did not work in our early experiments in the Sup. Mat.~\ref*{section:supp:alternative_latent_space_learning}.

\paragraph{Variational AutoEncoder (VAE)} The conventional latent diffusion pipeline~\cite{rombach2022stablediffusion} involves  learning a latent space with a Variational Autoencoder (VAE)~\cite{kingma2013auto}, as shown in Figure~\ref{fig:autoreconstructor}(a), where an encoder \( q_\phi \) takes a ground-truth sample \( X \) as input and maps it to a lower-dimensional latent distribution \( q_\phi(Z|X) \). A latent is sampled and decoded back to \( X' \) using a decoder \( p_\theta \). A loss is computed between \( X \) and \( X' \) to ensure faithful reconstruction of the data, and this loss is used to train both the encoder \( q_\phi \) and the decoder \( p_\theta \). However, the conventional latent space learning is not applicable in our case, because we do not have access to the ground-truth Splatter Images \( X \), but only to the images \( m = f(X) \), which are projections of \( X \).

\paragraph{Variational AutoReconstructor (VAR)}
To address this gap, we propose a new architecture called the Variational AutoReconstructor (VAR), as shown in Figure~\ref{fig:autoreconstructor} (b).
This encodes a 2D image \( m = f(X) \)—a projection of \( X \)—into a latent distribution \( q_\phi(Z|m) \), and then reconstructs a sample back to a high-dimensional representation \( X' \), which corresponds to the unknown 3D Splatter Image representation.
We supervise \( X' \) using posed multi-view images \( \{m_i\} \), each of which is a projection of \( X \), available in large-scale image datasets. The differentiable function \( f(\cdot) \) makes the entire pipeline end-to-end trainable.

\begin{figure*}[t]
    \includegraphics[width=\textwidth]{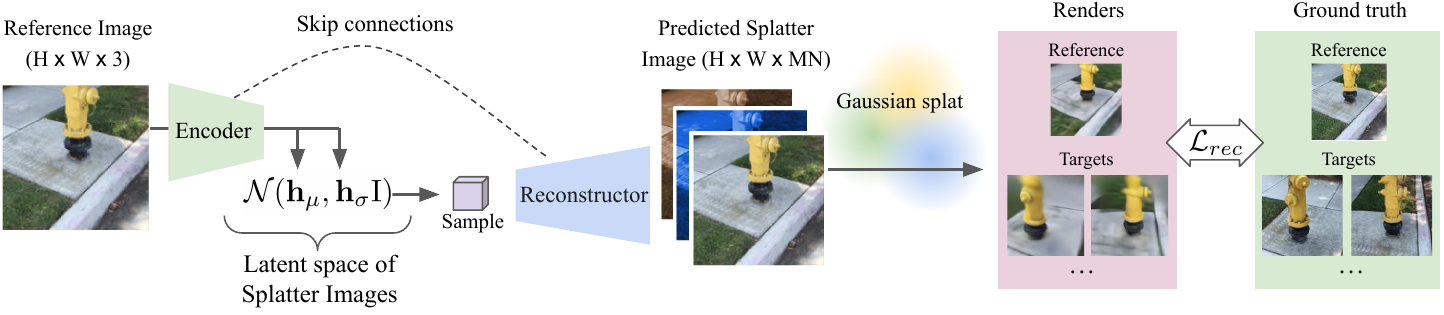}
    \vspace{-16pt}
    \caption[Learning a latent space for Splatter Images.]{
        \textbf{Learning a latent space for Splatter Images.} Our encoder predicts the parameters of a normal distribution over latents. We reconstruct a sampled latent into \( H \times W \times MN \) Splatter Image representations. We render the Gaussian splats from the viewpoints of the target training images and optimize reprojection losses between the rendered and ground-truth RGB images. Skip connections are critical to preserving the high-frequency details of the predictions, as shown in Figure~\ref{fig:skip-connection}.
    }
    \vspace{-5pt}
    \label{fig:training_step1}
\end{figure*}

Figure~\ref{fig:training_step1} gives an overview of the Variational AutoReconstructor training pipeline.
We assume access to training datasets containing multiple image sequences capturing different object instances or scenarios, such as CO3D~\cite{reizenstein21co3d} and RealEstate10K~\cite{zhou2018stereo}. Each training example comprises a reference image \( I^\text{ref} \), a set of target images \( \mathcal{I} = \{ I^\text{tgt}_{i} \mid i = 0, \ldots, t \} \), and the corresponding camera poses and intrinsics.
At training time, the reference image \( I^\text{ref} \) is input to the encoder, which produces a Gaussian distribution represented by a mean \( \code_{\mu} \) and variance \( \code_{\sigma} \). The latent code \( \code \) is sampled from the Gaussian distribution \( \mathcal{N}(\code_{\mu}, \code_{\sigma}\mathrm{I}) \) using the reparameterization trick~\cite{kingma2013auto}: 
\[
\code = \code_{\mu} + \epsilon \code_{\sigma}, \quad \text{where } \epsilon \sim \mathcal{N}(0, \mathrm{I}).
\]
This sampled latent code is passed to the reconstructor to obtain the Splatter Image representation corresponding to the reference view.
The predicted Splatter Image is then converted into a Gaussian splat representation of the scene and rendered into the target views using the known camera intrinsics and poses for each view $I_i$ in \( \mathcal{I} \), denoted as \( \hat{I_i} \). In practice, we also render the reference view.

\begin{figure}%
    \vspace{-2pt}
    \centering
    \includegraphics[width=0.92\columnwidth]{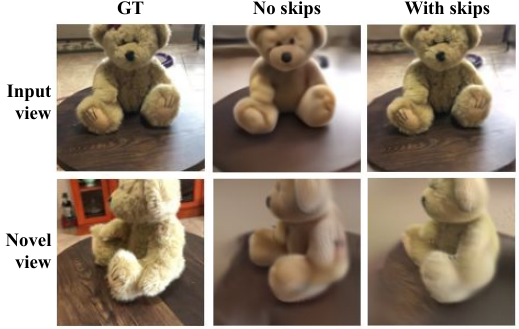}
    \caption[Skip connections.]{
    \textbf{Including skip connections} helps preserve high-frequency details from the input view in the AutoReconstructor, improving the faithfulness of appearance.
                }
    \vspace{-12pt}
    \label{fig:skip-connection}
\end{figure}

\paragraph{Preserving High-frequency Details with Skip connections}
We observe that a Splatter Image directly reconstructed from the low-dimensional latent space captures the geometry but loses high-frequency texture details. To overcome this, we introduce a skip connection~\cite{ronneberger2015u}, which allows high-frequency information to flow directly from the encoder's first-layer features to the higher-resolution layers of the reconstructor, as demonstrated in Figure~\ref{fig:skip-connection}. 
At test time, we gain flexibility by modifying the weight of skip connection features to control the faithfulness of the reconstruction to the input images, potentially balancing the diversity of reconstructions. 
More results are in the supplementary material section~\ref*{sec:sup_diverse_sampling}.

\paragraph{Losses}
The reconstruction loss $\mathcal{L}_{rec}$ for a single training example is:
\begin{equation}
\mathcal{L}_{rec} = \sum_{I \in \mathcal{I^{+}}} \lambda_1||I - \hat{I}||^2 + \lambda_2 \mathrm{SSIM}(I,\hat{I}) +
\lambda_3 \mathrm{LPIPS}(I,\hat{I})
\label{eq:rec}
\end{equation}
where $\mathcal{I^{+}} = \{I^{ref}\} \cup \mathcal{I}$ includes the reference and target images and $\lambda_{*}$ are weights for the different loss terms.
SSIM \cite{wang2004image} encourages similarities in structure and contrast. 
LPIPS~\cite{zhang2018perceptual} matches neural network encodings of image patches, which encourages photorealism. 
Following VAE~\cite{kingma2013auto}, we also use a KL divergence loss $\mathcal{L}_{KL}$ on the latent space \( z \) to regularize it to follow a prior Gaussian distribution:
\begin{equation}
\mathcal{L}_{KL} = \mathrm{KL}\left(\mathcal{N}(\code_{\mu}, \code_{\sigma}\mathrm{I}), \mathcal{N}(0, \mathrm{I})\right) .
\end{equation}

\subsection{Training the Latent Diffusion Model}

Once the Variational AutoReconstructor is trained, we encode all training images into latent code samples. A denoising diffusion model is then trained to learn the distribution of these latent codes conditioned on the input. The overall pipeline is illustrated in the supplement Fig.~\ref*{fig:training_step2}.

In general, we follow previous latent diffusion training pipelines~\cite{rombach2022stablediffusion, ke2023repurposing}.
The forward diffusion process for latent code $\code$ at time $t$ is 
\begin{equation}
\code_t = \sqrt{\bar{\alpha}_t}\code + \sqrt{1-\bar{\alpha}_t} \boldsymbol{\epsilon},
\end{equation}
~where $\boldsymbol{\epsilon}~\sim~\mathcal{N}(0, \mathrm{I})$ is the sampled Gaussian noise, $\bar{\alpha}_t = \prod_{s=1}^t{(1-\beta_s)}$ is the noise variance and $\beta_1, \beta_2, \dots, \beta_T$ is the variance schedule~\cite{ho2020denoising}.
During training we minimize the denoising loss:
\begin{equation}
    \mathbb{E}_{\code, \boldsymbol{\epsilon} \sim \mathcal{N}(0, \mathrm{I}), t \sim \mathcal{U}(T)} || \boldsymbol{\epsilon} - \boldsymbol{\epsilon}_\theta(\code_t, \phi^\text{ref}, t) ||_2 ^2 ,
\end{equation}
where $\boldsymbol{\epsilon}_\theta(\code_t, \phi^\text{ref}, t)$ is the denoising diffusion model that takes the noised latent code $\code_t$, input image features $\phi^\text{ref}$ and time step $t$ as input.

\paragraph{Conditioning Diffusion Models on Input Images}
Conventional diffusion models~\cite{rombach2022stablediffusion} take text prompts as conditions, which are encoded into feature vectors. Instead, we condition on the input reference image \( I^\text{ref} \), preserving the image structure and correspondence, to generate a Splatter Image.
Following Marigold~\cite{ke2023repurposing}, we concatenate features \( \phi^\text{ref} = F_{feat}(I^\text{ref}) \) extracted using a feature encoder $F_{feat}(.)$ from the input image \( I^\text{ref} \) with the noised latent codes as input to the denoiser.
There are several options for the feature encoder \( F_{\text{feat}}(\cdot) \). We directly use the pre-trained encoder from the Stable Diffusion VAE model~\cite{rombach2022stablediffusion} to obtain the conditional features \( \phi^\text{ref} \). As future work, we plan to explore other embeddings such as CLIP~\cite{radford2021learning} to condition on text prompts.

\paragraph{Controlling Diversity with Classifier-Free Guidance}
To increase the diversity of samples for generation tasks, we train with classifier-free guidance~\cite{rombach2022stablediffusion}, where conditioning features are zeroed out 20\% of the time (akin to unconditional generation). This reduces overfitting to the scene appearance visible in the conditioning view. During inference, it provides the option to increase the guidance weight to generate more diverse samples to balance between faithfulness and diversity.

\subsection{Diffusion Inference}

During inference, our method takes a single image as input and outputs a Gaussian splat represented as a Splatter Image (Fig.~\ref*{fig:inference}, in Sup.~Mat). The input image is first encoded by a feature encoder to generate conditioning features. A randomly initialized latent code is concatenated with these features and fed into the denoising diffusion model for 50 diffusion steps.
The resulting denoised latent code is then decoded into a Splatter Image using the Reconstructor trained in the first stage. This Splatter Image is subsequently backprojected to 3D to produce the final Gaussian splat.

\paragraph{Controlling Faithfulness and Diversity}
As a generative method, our model enables sampling multiple outputs by denoising different randomly initialized latent codes. 
The trade-off between fidelity to the input image and diversity of the generated results can be controlled by adjusting the weights of classifier-free guidance and skip connections.

\section{Experiments}

We introduce the basic experimental settings below and provide implementation details in the Supplementary Material.

\subsection{Experimental Settings}

\paragraph{Datasets}
We train and evaluate our model on the challenging CO3D dataset~\cite{reizenstein21co3d} and the RealEstate10K dataset~\cite{zhou2018stereo}. 
The CO3D dataset comprises 360-degree video captures of various objects in both indoor and outdoor real-world contexts, with annotated camera poses obtained via Structure-from-Motion (SfM)~\cite{schoenberger2016sfm}. 
Following prior works~\cite{szymanowicz2024splatter,wewer24latentsplat,mu2024gsd}, we train and evaluate on the most common categories, \textit{Hydrants} and \textit{Teddy Bears}, which include 723 and 1329 scenes, respectively.
The RealEstate10K dataset~\cite{zhou2018stereo} contains real-world videos of residential indoor and outdoor environments, with camera poses also derived using SfM. We use the provided official training and testing splits.

\paragraph{Baselines}  
We compare our method with several state-of-the-art baselines, including:
(1) Original \textit{Splatter Image}~\cite{szymanowicz2024splatter}, a forward method using Gaussian splatting, applied only on objects;
(2) \textit{Splatter Image (Full Images)}\(\star\), which we train on full images including the background;
(3) \textit{PixelNeRF}~\cite{yu2021pixelnerf}, a forward method based on NeRF;
(4) \textit{DFM}~\cite{tewari2023diffusion}, a diffusion-based NeRF model;
(5) \textit{ZeroNVS}~\cite{sargent2024zeronvs}, a novel view synthesis method;
(6) \textit{SparseFusion}~\cite{zhou2023sparsefusion}, which optimizes a scene using score distillation for each view; and
(7) \textit{LGM}~\cite{tang2024lgm}: A large-scale object reconstruction model based on Gaussian splats.
We introduce the baselines in detail in the supplementary material.

\paragraph{Metrics} 
Following prior work~\cite{yu2021pixelnerf,lin2023vision,szymanowicz2024splatter}, we report PSNR and LPIPS scores, where PSNR quantifies pixel-wise reconstruction fidelity and LPIPS measures perceptual similarity based on deep features between rendered and ground-truth images.  
Following~\cite{hollein2024viewdiff}, we report FID~\cite{heusel2017gans} and KID~\cite{binkowski2018demystifying} scores for generative performance, which measure image similarity at the distribution level by comparing statistics of deep image features over multiple images, thereby evaluating visual realism.  
We further report scores separately on \textit{Full Images}, which computes metrics on all pixels of reference views, and \textit{Objects Only}, which computes metrics only on object pixels following~\cite{szymanowicz2024splatter}, for a fair comparison with prior object-centric work.

\subsection{Reconstruction Performance}

\begin{table*}
    \resizebox{1.0\linewidth}{!}{
    \begin{tabular}{@{}clcccccccccc@{}}
            \toprule
            \textbf{Rendering}          & \multirow{2}{*}{\textbf{Methods}}                          & \multicolumn{2}{c}{\textbf{Time Cost / Sequence}}                                                             & \multicolumn{4}{c}{\textbf{Hydrants}}                                & \multicolumn{4}{c}{\textbf{Teddybears}}
            \\
            \cmidrule(l){3-12}
           \textbf{Speed} &  & \textbf{Inference} & \textbf{Scene Render}   & \textbf{PSNR ↑} & \textbf{LPIPS ↓} & \textbf{FID ↓} & \textbf{KID ↓} & \textbf{PSNR ↑} & \textbf{LPIPS ↓} & \textbf{FID ↓} & \textbf{KID ↓}
           \\
           \midrule
            \multirow{2}{*}{Offline}   & PixelNeRF \cite{yu2021pixelnerf} $\dagger$ & 5.0 sec & 4.1 min                                                         & \textbf{17.93}  & 0.54            & 180.20          & 0.14          & -               & -                & -              & -              \\
                                       & DFM \cite{tewari2023diffusion} $\dagger$ & 2 min & 8 min                                                             & 17.47           & \textbf{0.42}   & \textbf{84.63}  & \textbf{0.05} & -               & -                & -              & 
            \\
            \midrule
             & Splatter Image (Full Images) $\star$                       & 49.4 ms & 109.5 ms                                                        & 17.37           & 0.492            & 155.4         & 0.127          & {\ul 17.44}     & {\ul 0.478}            & 102.8         & 0.069          
            \\
           \rowcolor[HTML]{EFEFEF} &  Ours - AutoReconstructor $\star$                                & \textbf{15.7 ms} &   \textbf{29.1 ms}   & {\ul 17.59}           & {\ul 0.471}            & {\ul 133.0}         & \underline{0.111}    & \textbf{17.49}  & {\bf 0.477}      & \textbf{95.1} & \textbf{0.058} 
           \\
           \rowcolor[HTML]{EFEFEF} &  Ours - Diffusion single sample $\star$                       & 2.7 sec & \textbf{29.1 ms} & 17.40           & 0.473            & 133.8         & { 0.114}          & 16.77           & 0.480            & {\ul 97.8}          & {\ul 0.061}
           
           \\ 
           \cmidrule(l){2-12}
           \cmidrule(l){2-12}
           
           \rowcolor[HTML]{EFEFEF} \multirow{-4}{*}{Real-time} &  Ours - Diffusion 20-best oracle $\star$                      & 54.0 sec &          {29.1 ms}                                                                & {17.74}     &  {0.466}      & {132.3}   & { 0.114}          & 17.08           & {0.474}   & { 96.6}    & 0.062
           
           \\
           \bottomrule
    \end{tabular}
    }
    \vspace{-5pt}
    \caption[Quantitative results on full images from CO3D.]{
        \textbf{Quantitative results on full images from CO3D.}  We compare with other methods that also perform single-image-input novel view synthesis on full images, rather than only on masked objects. Our model outperforms other real-time rendering baselines. $\dagger$: cited from \cite{tewari2023diffusion}; $\star$: trained by us. The reported inference times account for the complete pipeline.
    }
    \label{tab:whole_object_results}
    \vspace{-6pt}
\end{table*}

\begin{figure}
    \centering
    \includegraphics[width=1.0\columnwidth]{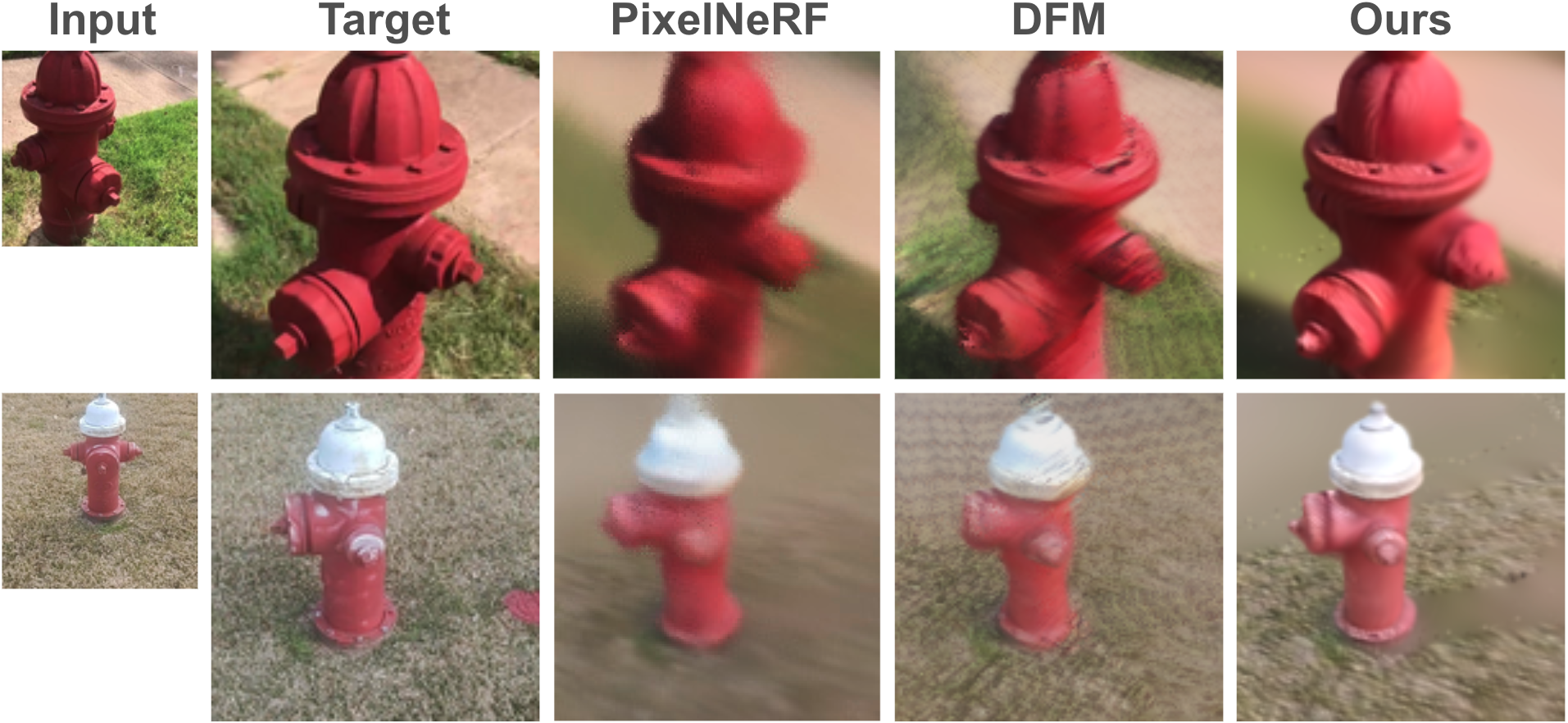}
    \caption{
    \textbf{Qualitative results} on the ``Hydrants'' category from the CO3D dataset. We produce significantly sharper results than PixelNeRF~\cite{yu2021pixelnerf}, and comparable or better performance on object areas compared to DFM~\cite{tewari2023diffusion}, while being significantly faster. Computation times are reported in Table~\ref{tab:whole_object_results}.
    }
    \label{fig:vs_pixelnerf}
    \vspace{-10pt}
\end{figure}

\begin{figure}
    \includegraphics[width=\columnwidth]{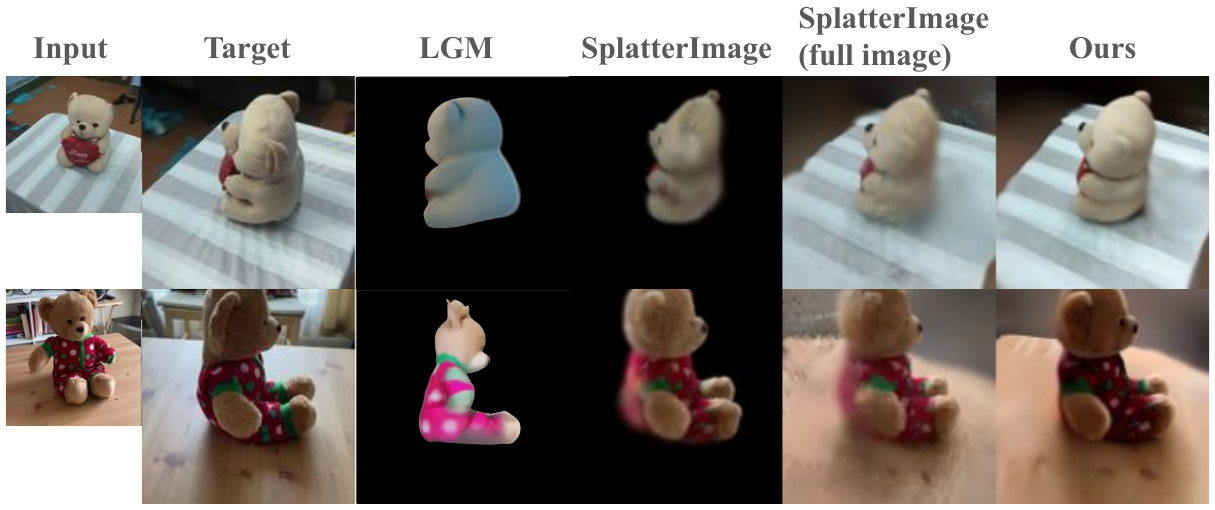}
    \vspace{-8pt}
    \caption{
        \textbf{Qualitative results} on the ``TeddyBears'' category from the CO3D dataset. Our model produces sharper results with higher-quality details compared to the baselines LGM~\cite{tang2024lgm} and SplatterImage~\cite{szymanowicz2024splatter}, especially in occluded areas.
    }
    \label{fig:results_teddybears}
    \vspace{-10pt}
\end{figure}

\paragraph{CO3D Datasets}
We present quantitative results on full images in Table~\ref{tab:whole_object_results}.  
The results show that our method outperforms real-time baselines on most metrics.  
Although DFM~\cite{tewari2023diffusion} outperforms our method on some metrics, it is considerably slower due to its NeRF representation and per-frame diffusion design.  
PixelNeRF~\cite{yu2021pixelnerf} achieves slightly higher PSNR than our method but performs significantly worse in LPIPS, likely because it produces generic smooth details in unobserved regions. 
As shown in Fig.~\ref{fig:vs_pixelnerf} and Fig.~\ref{fig:results_teddybears}, our results appear considerably sharper than those of PixelNeRF and SplatterImage, and are comparable to DFM, while being much faster to compute (3 seconds versus 10 minutes). 
We output real-world scenes with both foreground and background, in comparison to object-only models such as LGM~\cite{tang2024lgm}.
We provide a detailed quantitative evaluation of object reconstruction methods in Table~\ref*{tab:masked_object_results} 
in the Supplementary Materials.

\begin{figure}
    \centering
    \includegraphics[width=1.0\columnwidth]{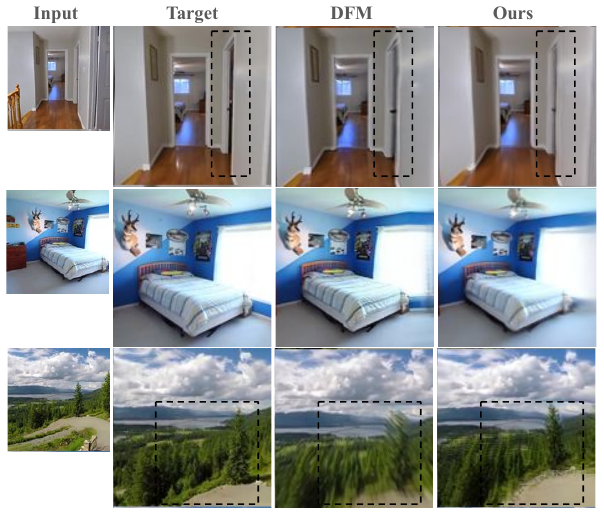}
    \vspace{-15pt}
    \caption[Qualitative Results on the RealEstate10K Dataset.]{\textbf{Qualitative Results on the RealEstate10K Dataset.}  
Our method achieves comparable performance to DFM~\cite{tewari2023diffusion}, a diffusion-based NeRF model, and performs better in some challenging regions (highlighted with dotted boxes), while being significantly faster at inference time.

    }
    \label{fig:re10k_qualitative}
    \vspace{-10pt}
\end{figure}

\paragraph{RealEstate10K Datasets}
We present quantitative results on the RealEstate10K dataset in Table~\ref{table:realestate10k}.  
To fairly compare with baselines that follow different evaluation settings, we follow PixelNeRF~\cite{yu2021pixelnerf}, SparseFusion~\cite{zhou2023sparsefusion}, and DFM~\cite{tewari2023diffusion} by evaluating on 100 scenes at a resolution of $128 \times 128$.  
We also follow ZeroNVS~\cite{sargent2024zeronvs} to evaluate on all 6,473 scenes and report results at both $128 \times 128$ resolution and $256 \times 256$, with the latter obtained by upsampling our rendered images.  
Our method outperforms SparseFusion, PixelNeRF, and ZeroNVS by a significant margin, as we generate a consistent 3D representation. In contrast, SparseFusion and ZeroNVS suffer from view inconsistency due to their reliance on multi-view optimization, and the intensive computation. 
We also achieve performance comparable to DFM while being significantly faster at inference time. 
We show qualitative results compared to DFM in Figure~\ref{fig:re10k_qualitative}, where we achieve better performance in some challenging regions.

\begin{table*}
\centering
\resizebox{0.75\textwidth}{!}{
\begin{tabular}{l|c|cc|cccc}
\hline
\textbf{Methods}                  & \textbf{Rendering speed}                       & \textbf{\# Scenes}         & \textbf{Resolution}                         & \textbf{PSNR ↑}                       & \textbf{LPIPS ↓}                       & \textbf{FID ↓}                         & \textbf{KID ↓}                        \\ \hline
pixelNeRF~\cite{yu2021pixelnerf} $\dagger$  &  Offline       &              100           &                  128                    & -                                     & -                                      & 195.4                                  & 0.14                                  \\
SparseFusion~\cite{zhou2023sparsefusion} $\dagger$ &  Offline  &            100             &                128                      & -                                     & -                                      & 99.44                                  & 0.04                                  \\
DFM~\cite{tewari2023diffusion}   &  Offline          &           100              &                128                      & -                                     & -                                      & \textbf{42.84}                                  & \textbf{0.01}                                  \\
\cellcolor[HTML]{EFEFEF}Ours Diffusion   &  \cellcolor[HTML]{EFEFEF}Real-time                         & \cellcolor[HTML]{EFEFEF}100   & \cellcolor[HTML]{EFEFEF}128                & \cellcolor[HTML]{EFEFEF}19.9 & \cellcolor[HTML]{EFEFEF}0.238 & \cellcolor[HTML]{EFEFEF}{\ul 49.83}& \cellcolor[HTML]{EFEFEF}\textbf{0.01} \\ \hline
ZeroNVS~\cite{sargent2024zeronvs}   &  Offline       &            6,473             & 256                                  & 13.5                                  & 0.414                                  & -                                      & -                                     \\
\cellcolor[HTML]{EFEFEF}Ours Diffusion    &  \cellcolor[HTML]{EFEFEF}Real-time                                &         \cellcolor[HTML]{EFEFEF} 6,473               & \cellcolor[HTML]{EFEFEF}128          & \cellcolor[HTML]{EFEFEF}\textbf{20.2}          & \cellcolor[HTML]{EFEFEF}\textbf{0.253}          & \cellcolor[HTML]{EFEFEF}\textbf{15.53}          & \cellcolor[HTML]{EFEFEF}\textbf{0.01}          \\
\cellcolor[HTML]{EFEFEF}Ours Diffusion     &  \cellcolor[HTML]{EFEFEF}Real-time                      & \cellcolor[HTML]{EFEFEF} 6,473 & \cellcolor[HTML]{EFEFEF}256 & \cellcolor[HTML]{EFEFEF}{\ul 19.6} & \cellcolor[HTML]{EFEFEF}{\ul 0.365} & \cellcolor[HTML]{EFEFEF}33.44 & \cellcolor[HTML]{EFEFEF}0.03 \\ \hline
\end{tabular}
}
\vspace{-5pt}
\caption[RealEstate10K Results.]{\textbf{RealEstate10K Results.} We report the number of test scenes and resolution following the settings used in prior work to ensure fair comparison.  
The baseline results are cited from the corresponding published papers. 
$\dagger$: cited from \cite{tewari2023diffusion};
``--'': Results are not reported.
 }
\label{table:realestate10k}
\vspace{-12pt}
\end{table*}

\subsection{Generative Performance}

Compared with regression-based baselines, our method can generate diverse samples with varying color, structure, and style for unobserved and uncertain parts, as shown in Figure~\ref{fig:gaussian_title_figure}. We demonstrate this ability in detail below.

\begin{figure}
    \centering
    \includegraphics[width=\columnwidth]{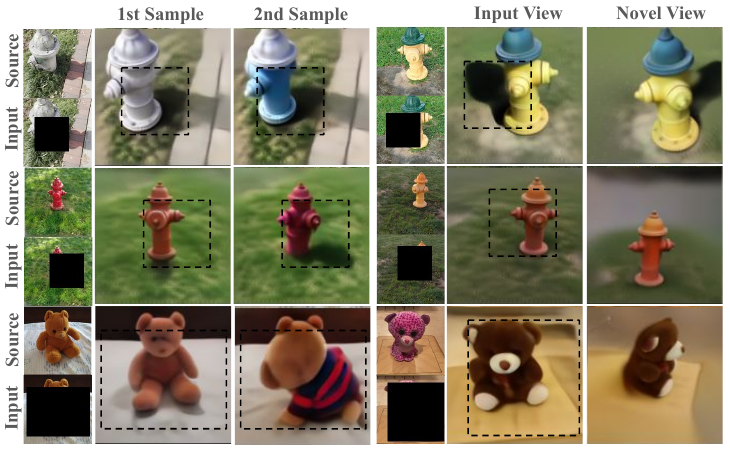}
    \vspace{-10pt}
    \caption{
        \textbf{3D Generative Performance.} Our diffusion model demonstrates the ability to (1) sample diverse outputs in ambiguous situations, and (2) fill in missing areas using 3D priors learned from large datasets with multi-view consistency. Note the model is trained purely from only 2D images.
    }
    \label{fig:results_3d_generation}
    \vspace{-10pt}
\end{figure}

\paragraph{Qualitative Results}
As shown in Figure~\ref{fig:results_3d_generation}, our diffusion model is able to sample diverse results in ambiguous situations where parts of objects are occluded with black boxes. The filled-in areas exhibit reasonable 3D structure consistent with the visible parts and remain consistent as part of a holistic 3D representation when viewed from novel perspectives.  
This demonstrates that the model effectively learns 3D priors purely from 2D images.
We further demonstrate this in Figure~\ref*{fig:diffusion_samples} in the Supplementary Materials, where, compared with DFM, we are able to control the diversity in the occluded back of objects, showing increasingly strong differences to balance fidelity to the input image against output diversity.

\paragraph{Quantitative: \(k\)-best oracle evaluation} 
Generative models are capable of producing multiple possible predictions. Oracle $k$-best evaluation~\cite{henderson2024sampling,guzman2012multiple,bui20206d} allows the model to produce \(k=20\) samples for each input, and the score of the best sample is used to evaluate that input image. 
This metric shows an upper bound reconstruction performance, assuming some manual effort can be made to select the best result. It also reflects the generative model's ability to capture the distribution and produce diverse samples.
The results of our diffusion model on the CO3D datasets are shown in Table~\ref{tab:whole_object_results} and the Sup. Mat.. 
We observe an improvement in the reported metrics compared to single samples, which demonstrates that our model generates diverse outputs rather than collapsing to a single mode.

\subsection{Ablation Studies}
We ablate the key modules and hyperparameters of our AutoReconstructor to validate our contributions in Table~\ref{tab:ablations}. 
Our method scores highest or second highest across all metrics, validating our design decisions. The ``No Augmentation'' row shows that our augmentations boost PSNR and LPIPS at the expense of a small drop in FID and KID scores.
More discussions are given in the Supplementary Materials.

\begin{table}
    \centering
    \resizebox{0.95\columnwidth}{!}{
            \begin{tabular}{@{}lcccc@{}}
            \toprule
            \textbf{Encoder-Reconstructor Variants} & \textbf{PSNR ↑} & \textbf{LPIPS ↓} & \textbf{FID ↓} & \textbf{KID ↓} \\ \midrule
            Splatter Image                    & 17.37          & 0.492            & 155.4          & 0.127          \\ \hline
            No skip connections               & 16.63          & 0.551            & 167.9          & 0.143          \\
            Skips from first 2 layers         & 16.86          & 0.554            & 163.3          & 0.135          \\
            One Gaussian splat per pixel      & 17.43          & 0.491            & 140.1          & 0.115          \\
            No variational sampling           & 8.30           & 0.726            & 290.1          & 0.293          \\
            No SSIM                           & 17.34          & 0.505            & 181.1          & 0.162          \\
            No LPIPS                          & {\ul 17.51}    & 0.502            & 149.7          & 0.118          \\
            No Color Augmentation                   & 17.20          & {\ul 0.486}      & \textbf{132.4} & \textbf{0.108} \\
            \cellcolor[HTML]{EFEFEF} Our AutoReconstructor               & \cellcolor[HTML]{EFEFEF}\textbf{17.59} & \cellcolor[HTML]{EFEFEF}\textbf{0.471}   & \cellcolor[HTML]{EFEFEF}{\ul 133.0}    & \cellcolor[HTML]{EFEFEF}{\ul 0.111}    \\ \bottomrule
            \end{tabular}
    }
    \caption[Ablation studies on the Hydrants scenes.]{\textbf{Ablation studies} on the Hydrants scenes from the CO3D dataset evaluated on \emph{Full Images}.}
    \label{tab:ablations}
    \vspace{-10pt}
\end{table}

\section{Conclusion}

We presented a training pipeline for a denoising diffusion model that reconstructs complete 3D scenes using Gaussian splats from a single RGB image as input.  
By proposing a Variational AutoReconstructor architecture, we are able to efficiently learn a latent space for Splatter Images using only 2D images, from which a denoising diffusion model can be trained by conditioning on input image features.
Experiments on the CO3D and RealEstate10K datasets demonstrate that our approach achieves state-of-the-art results in scene reconstruction, particularly in occluded regions.  
Furthermore, our method enables diverse sampling of Gaussian splats with controllable trade-offs between faithfulness to the input and generative diversity.

\clearpage
{
    \small
    \bibliographystyle{ieeenat_fullname}
    \bibliography{main}

\begin{thebibliography}{84}
\providecommand{\natexlab}[1]{#1}
\providecommand{\url}[1]{\texttt{#1}}
\expandafter\ifx\csname urlstyle\endcsname\relax
  \providecommand{\doi}[1]{doi: #1}\else
  \providecommand{\doi}{doi: \begingroup \urlstyle{rm}\Url}\fi

\bibitem[Bi{\'n}kowski et~al.(2018)Bi{\'n}kowski, Sutherland, Arbel, and Gretton]{binkowski2018demystifying}
Miko{\l}aj Bi{\'n}kowski, Danica~J Sutherland, Michael Arbel, and Arthur Gretton.
\newblock Demystifying {MMD GANs}.
\newblock In \emph{ICLR}, 2018.

\bibitem[Boss et~al.(2024)Boss, Huang, Vasishta, and Jampani]{boss2024sf3d}
Mark Boss, Zixuan Huang, Aaryaman Vasishta, and Varun Jampani.
\newblock {SF3D}: Stable fast {3D} mesh reconstruction with uv-unwrapping and illumination disentanglement.
\newblock In \emph{{CVPR}}, 2024.

\bibitem[Bui et~al.(2020)Bui, Birdal, Deng, Albarqouni, Guibas, Ilic, and Navab]{bui20206d}
Mai Bui, Tolga Birdal, Haowen Deng, Shadi Albarqouni, Leonidas Guibas, Slobodan Ilic, and Nassir Navab.
\newblock {6D} camera relocalization in ambiguous scenes via continuous multimodal inference.
\newblock In \emph{ECCV}, 2020.

\bibitem[Chan et~al.(2021)Chan, Monteiro, Kellnhofer, Wu, and Wetzstein]{chanmonteiro2020piGAN}
Eric Chan, Marco Monteiro, Petr Kellnhofer, Jiajun Wu, and Gordon Wetzstein.
\newblock {pi-GAN}: Periodic implicit generative adversarial networks for {3D}-aware image synthesis.
\newblock In \emph{CVPR}, 2021.

\bibitem[Chan et~al.(2022)Chan, Lin, Chan, Nagano, Pan, Mello, Gallo, Guibas, Tremblay, Khamis, Karras, and Wetzstein]{Chan2021}
Eric~R. Chan, Connor~Z. Lin, Matthew~A. Chan, Koki Nagano, Boxiao Pan, Shalini~De Mello, Orazio Gallo, Leonidas Guibas, Jonathan Tremblay, Sameh Khamis, Tero Karras, and Gordon Wetzstein.
\newblock Efficient geometry-aware {3D} generative adversarial networks.
\newblock In \emph{CVPR}, 2022.

\bibitem[Chan et~al.(2023)Chan, Nagano, Chan, Bergman, Park, Levy, Aittala, Mello, Karras, and Wetzstein]{chan2023genvs}
Eric~R. Chan, Koki Nagano, Matthew~A. Chan, Alexander~W. Bergman, Jeong~Joon Park, Axel Levy, Miika Aittala, Shalini~De Mello, Tero Karras, and Gordon Wetzstein.
\newblock {GeNVS}: Generative novel view synthesis with {3D}-aware diffusion models.
\newblock In \emph{arXiv}, 2023.

\bibitem[Charatan et~al.(2024)Charatan, Li, Tagliasacchi, and Sitzmann]{charatan2024pixelsplat}
David Charatan, Sizhe~Lester Li, Andrea Tagliasacchi, and Vincent Sitzmann.
\newblock {pixelSplat}: {3D} gaussian splats from image pairs for scalable generalizable {3D} reconstruction.
\newblock In \emph{{CVPR}}, 2024.

\bibitem[Chen et~al.(2024{\natexlab{a}})Chen, Xu, Zheng, Zhuang, Pollefeys, Geiger, Cham, and Cai]{hen2024mvsplat}
Yuedong Chen, Haofei Xu, Chuanxia Zheng, Bohan Zhuang, Marc Pollefeys, Andreas Geiger, Tat-Jen Cham, and Jianfei Cai.
\newblock {MVSplat}: Efficient {3D} gaussian splatting from sparse multi-view images.
\newblock In \emph{ECCV}, 2024{\natexlab{a}}.

\bibitem[Chen et~al.(2024{\natexlab{b}})Chen, Wang, Wang, and Liu]{chen2024text}
Zilong Chen, Feng Wang, Yikai Wang, and Huaping Liu.
\newblock Text-to-{3D} using gaussian splatting.
\newblock In \emph{CVPR}, 2024{\natexlab{b}}.

\bibitem[Dai et~al.(2017{\natexlab{a}})Dai, Chang, Savva, Halber, Funkhouser, and Nie{\ss}ner]{dai2017scannet}
Angela Dai, Angel~X Chang, Manolis Savva, Maciej Halber, Thomas Funkhouser, and Matthias Nie{\ss}ner.
\newblock Scannet: Richly-annotated 3d reconstructions of indoor scenes.
\newblock In \emph{CVPR}, 2017{\natexlab{a}}.

\bibitem[Dai et~al.(2017{\natexlab{b}})Dai, Ruizhongtai~Qi, and Nie{\ss}ner]{dai2017shape}
Angela Dai, Charles Ruizhongtai~Qi, and Matthias Nie{\ss}ner.
\newblock Shape completion using {3D}-encoder-predictor cnns and shape synthesis.
\newblock In \emph{{CVPR}}, 2017{\natexlab{b}}.

\bibitem[Dai et~al.(2018)Dai, Ritchie, Bokeloh, Reed, Sturm, and Nie{\ss}ner]{dai2018scancomplete}
Angela Dai, Daniel Ritchie, Martin Bokeloh, Scott Reed, J{\"u}rgen Sturm, and Matthias Nie{\ss}ner.
\newblock {ScanComplete}: Large-scale scene completion and semantic segmentation for {3D} scans.
\newblock In \emph{{CVPR}}, 2018.

\bibitem[Dai et~al.(2020)Dai, Diller, and Nie{\ss}ner]{dai2020sg}
Angela Dai, Christian Diller, and Matthias Nie{\ss}ner.
\newblock {SG-NN}: Sparse generative neural networks for self-supervised scene completion of {RGB-D} scans.
\newblock In \emph{{CVPR}}, 2020.

\bibitem[Deitke et~al.(2023)Deitke, Schwenk, Salvador, Weihs, Michel, VanderBilt, Schmidt, Ehsani, Kembhavi, and Farhadi]{deitke2023objaverse}
Matt Deitke, Dustin Schwenk, Jordi Salvador, Luca Weihs, Oscar Michel, Eli VanderBilt, Ludwig Schmidt, Kiana Ehsani, Aniruddha Kembhavi, and Ali Farhadi.
\newblock Objaverse: A universe of annotated {3D} objects.
\newblock In \emph{CVPR}, 2023.

\bibitem[Fei et~al.(2024)Fei, Zheng, Duan, Zhan, Tomizuka, Keutzer, and Lu]{fei2024pixelgaussian}
Xin Fei, Wenzhao Zheng, Yueqi Duan, Wei Zhan, Masayoshi Tomizuka, Kurt Keutzer, and Jiwen Lu.
\newblock {PixelGaussian:} generalizable 3d gaussian reconstruction from arbitrary views.
\newblock \emph{arXiv preprint arXiv:2410.18979}, 2024.

\bibitem[Feng et~al.(2024)Feng, Xing, Wu, and Jiang]{feng2024fdgaussian}
Qijun Feng, Zhen Xing, Zuxuan Wu, and Yu-Gang Jiang.
\newblock {FDGaussian}: Fast gaussian splatting from single image via geometric-aware diffusion model.
\newblock \emph{arXiv}, 2024.

\bibitem[Galvis et~al.(2024)Galvis, Zuo, Schaefer, and Leutengger]{galvis2024sc}
Juan~D Galvis, Xingxing Zuo, Simon Schaefer, and Stefan Leutengger.
\newblock {SC-Diff: 3D} shape completion with latent diffusion models.
\newblock \emph{arXiv}, 2024.

\bibitem[Gao et~al.(2022)Gao, Shen, Wang, Chen, Yin, Li, Litany, Gojcic, and Fidler]{gao2022get3d}
Jun Gao, Tianchang Shen, Zian Wang, Wenzheng Chen, Kangxue Yin, Daiqing Li, Or Litany, Zan Gojcic, and Sanja Fidler.
\newblock {GET3D}: A generative model of high quality {3D} textured shapes learned from images.
\newblock In \emph{NeurIPS}, 2022.

\bibitem[Gao et~al.(2024)Gao, Holynski, Henzler, Brussee, Martin-Brualla, Srinivasan, Barron, and Poole]{gao2024cat3d}
Ruiqi Gao, Aleksander Holynski, Philipp Henzler, Arthur Brussee, Ricardo Martin-Brualla, Pratul Srinivasan, Jonathan~T Barron, and Ben Poole.
\newblock {Cat3D}: Create anything in {3D} with multi-view diffusion models.
\newblock In \emph{NeurIPS}, 2024.

\bibitem[Gu et~al.(2023)Gu, Trevithick, Lin, Susskind, Theobalt, Liu, and Ramamoorthi]{gu2023nerfdiff}
Jiatao Gu, Alex Trevithick, Kai-En Lin, Joshua~M Susskind, Christian Theobalt, Lingjie Liu, and Ravi Ramamoorthi.
\newblock {NerfDiff}: Single-image view synthesis with nerf-guided distillation from {3D}-aware diffusion.
\newblock In \emph{ICML}, 2023.

\bibitem[Guzman-Rivera et~al.(2012)Guzman-Rivera, Batra, and Kohli]{guzman2012multiple}
Abner Guzman-Rivera, Dhruv Batra, and Pushmeet Kohli.
\newblock Multiple choice learning: Learning to produce multiple structured outputs.
\newblock \emph{NeurIPS}, 25, 2012.

\bibitem[He et~al.(2025)He, Xu, Guo, Wetzstein, Dai, Li, and Yang]{he2025cameractrl}
Hao He, Yinghao Xu, Yuwei Guo, Gordon Wetzstein, Bo Dai, Hongsheng Li, and Ceyuan Yang.
\newblock Camera{C}trl: Enabling camera control for video diffusion models.
\newblock In \emph{ICLR}, 2025.

\bibitem[Henderson et~al.(2024)Henderson, de~Almeida, Ivanova, and Anciukevi{\v{c}}ius]{henderson2024sampling}
Paul Henderson, Melonie de Almeida, Daniela Ivanova, and Titas Anciukevi{\v{c}}ius.
\newblock Sampling {3D} gaussian scenes in seconds with latent diffusion models.
\newblock \emph{arXiv}, 2024.

\bibitem[Heusel et~al.(2017)Heusel, Ramsauer, Unterthiner, Nessler, and Hochreiter]{heusel2017gans}
Martin Heusel, Hubert Ramsauer, Thomas Unterthiner, Bernhard Nessler, and Sepp Hochreiter.
\newblock {GANs} trained by a two time-scale update rule converge to a local nash equilibrium.
\newblock \emph{NeurIPS}, 2017.

\bibitem[Ho et~al.(2020)Ho, Jain, and Abbeel]{ho2020denoising}
Jonathan Ho, Ajay Jain, and Pieter Abbeel.
\newblock Denoising diffusion probabilistic models.
\newblock \emph{NeurIPS}, 2020.

\bibitem[Ho et~al.(2022)Ho, Chan, Saharia, Whang, Gao, Gritsenko, Kingma, Poole, Norouzi, Fleet, and Salimans]{ho2022imagen}
Jonathan Ho, William Chan, Chitwan Saharia, Jay Whang, Ruiqi Gao, Alexey Gritsenko, Diederik~P Kingma, Ben Poole, Mohammad Norouzi, David~J Fleet, and Tim Salimans.
\newblock Imagen video: High definition video generation with diffusion models.
\newblock \emph{arXiv preprint arXiv:2210.02303}, 2022.

\bibitem[H{\"o}llein et~al.(2024)H{\"o}llein, Bo{\v{z}}i{\v{c}}, M{\"u}ller, Novotny, Tseng, Richardt, Zollh{\"o}fer, and Nie{\ss}ner]{hollein2024viewdiff}
Lukas H{\"o}llein, Alja{\v{z}} Bo{\v{z}}i{\v{c}}, Norman M{\"u}ller, David Novotny, Hung-Yu Tseng, Christian Richardt, Michael Zollh{\"o}fer, and Matthias Nie{\ss}ner.
\newblock {ViewDiff}: {3D}-consistent image generation with text-to-image models.
\newblock In \emph{{CVPR}}, 2024.

\bibitem[Hong et~al.(2024)Hong, Jung, Shin, Han, Yang, Luo, and Kim]{hong2024pf3plat}
Sunghwan Hong, Jaewoo Jung, Heeseong Shin, Jisang Han, Jiaolong Yang, Chong Luo, and Seungryong Kim.
\newblock {PF3plat}: Pose-free feed-forward 3d gaussian splatting.
\newblock \emph{arXiv preprint arXiv:2410.22128}, 2024.

\bibitem[Huang et~al.(2024)Huang, Stojanov, Thai, Jampani, and Rehg]{huang2024zeroshape}
Zixuan Huang, Stefan Stojanov, Anh Thai, Varun Jampani, and James~M Rehg.
\newblock {ZeroShape}: Regression-based zero-shot shape reconstruction.
\newblock In \emph{{CVPR}}, 2024.

\bibitem[Jang and Agapito(2024)]{jang2024nvist}
Wonbong Jang and Lourdes Agapito.
\newblock {NViST}: In the wild new view synthesis from a single image with transformers.
\newblock In \emph{{CVPR}}, 2024.

\bibitem[Jun and Nichol(2023)]{jun2023shap}
Heewoo Jun and Alex Nichol.
\newblock Shap-{E}: Generating conditional {3D} implicit functions.
\newblock \emph{arXiv}, 2023.

\bibitem[Karnewar et~al.(2023)Karnewar, Vedaldi, Novotny, and Mitra]{karnewar2023holodiffusion}
Animesh Karnewar, Andrea Vedaldi, David Novotny, and Niloy Mitra.
\newblock {HoloDiffusion}: Training a {3D} diffusion model using {2D} images.
\newblock In \emph{CVPR}, 2023.

\bibitem[Ke et~al.(2024)Ke, Obukhov, Huang, Metzger, Daudt, and Schindler]{ke2023repurposing}
Bingxin Ke, Anton Obukhov, Shengyu Huang, Nando Metzger, Rodrigo~Caye Daudt, and Konrad Schindler.
\newblock Repurposing diffusion-based image generators for monocular depth estimation.
\newblock In \emph{{CVPR}}, 2024.

\bibitem[Kerbl et~al.(2023)Kerbl, Kopanas, Leimk{\"u}hler, and Drettakis]{kerbl3Dgaussians}
Bernhard Kerbl, Georgios Kopanas, Thomas Leimk{\"u}hler, and George Drettakis.
\newblock {3D Gaussian Splatting for Real-Time Radiance Field Rendering}.
\newblock \emph{ToG}, 2023.

\bibitem[Kingma and Welling(2014)]{kingma2013auto}
Diederik~P Kingma and Max Welling.
\newblock Auto-encoding variational bayes.
\newblock In \emph{ICLR}, 2014.

\bibitem[Li et~al.(2023)Li, Wang, and Tseng]{li2023gaussiandiffusion}
Xinhai Li, Huaibin Wang, and Kuo-Kun Tseng.
\newblock {GaussianDiffusion}: {3D} gaussian splatting for denoising diffusion probabilistic models with structured noise.
\newblock \emph{arXiv}, 2023.

\bibitem[Liang et~al.(2025)Liang, Cao, Goel, Qian, Korolev, Terzopoulos, Plataniotis, Tulyakov, and Ren]{liang2025wonderland}
Hanwen Liang, Junli Cao, Vidit Goel, Guocheng Qian, Sergei Korolev, Demetri Terzopoulos, Konstantinos~N Plataniotis, Sergey Tulyakov, and Jian Ren.
\newblock Wonderland: Navigating 3d scenes from a single image.
\newblock In \emph{CVPR}, 2025.

\bibitem[Liao et~al.(2024)Liao, Xu, and Waslander]{liao2024toward}
Ziwei Liao, Binbin Xu, and Steven~L Waslander.
\newblock Toward general object-level mapping from sparse views with {3D} diffusion priors.
\newblock In \emph{CoRL}, 2024.

\bibitem[Lin et~al.(2025)Lin, Pan, Yang, Li, and Mu]{lin2025diffsplat}
Chenguo Lin, Panwang Pan, Bangbang Yang, Zeming Li, and Yadong Mu.
\newblock Diffsplat: Repurposing image diffusion models for scalable gaussian splat generation.
\newblock In \emph{{ICLR}}, 2025.

\bibitem[Lin et~al.(2023)Lin, Lin, Lai, Lin, Shih, and Ramamoorthi]{lin2023vision}
Kai-En Lin, Yen-Chen Lin, Wei-Sheng Lai, Tsung-Yi Lin, Yi-Chang Shih, and Ravi Ramamoorthi.
\newblock Vision transformer for nerf-based view synthesis from a single input image.
\newblock In \emph{{WACV}}, 2023.

\bibitem[Liu et~al.(2024)Liu, Xu, Cheng, Gao, Wang, Shan, and Tang]{liu2024novelgs}
Jinpeng Liu, Jiale Xu, Weihao Cheng, Yiming Gao, Xintao Wang, Ying Shan, and Yansong Tang.
\newblock {NovelGS}: Consistent novel-view denoising via large gaussian reconstruction model.
\newblock \emph{arXiv preprint arXiv:2411.16779}, 2024.

\bibitem[Liu et~al.(2023)Liu, Wu, Van~Hoorick, Tokmakov, Zakharov, and Vondrick]{liu2023zero}
Ruoshi Liu, Rundi Wu, Basile Van~Hoorick, Pavel Tokmakov, Sergey Zakharov, and Carl Vondrick.
\newblock Zero-1-to-3: Zero-shot one image to {3D} object.
\newblock In \emph{ICCV}, 2023.

\bibitem[Liu et~al.(2025)Liu, Fan, Yu, Li, Lu, and Yuan]{liu2025monosplat}
Yifan Liu, Keyu Fan, Weihao Yu, Chenxin Li, Hao Lu, and Yixuan Yuan.
\newblock {MonoSplat}: Generalizable 3d gaussian splatting from monocular depth foundation models.
\newblock In \emph{CVPR}, 2025.

\bibitem[Ma et~al.(2025)Ma, Gao, Deng, Luo, Huang, Tang, and Wang]{ma2025you}
Baorui Ma, Huachen Gao, Haoge Deng, Zhengxiong Luo, Tiejun Huang, Lulu Tang, and Xinlong Wang.
\newblock You see it, you got it: Learning 3d creation on pose-free videos at scale.
\newblock In \emph{CVPR}, pages 2016--2029, 2025.

\bibitem[Meng et~al.(2025)Meng, Wang, Lei, Daniilidis, Gu, and Liu]{meng2025zero}
Xuyi Meng, Chen Wang, Jiahui Lei, Kostas Daniilidis, Jiatao Gu, and Lingjie Liu.
\newblock {Zero-1-to-G}: Taming pretrained 2d diffusion model for direct 3d generation.
\newblock \emph{arXiv preprint arXiv:2501.05427}, 2025.

\bibitem[Mildenhall et~al.(2020)Mildenhall, Srinivasan, Tancik, Barron, Ramamoorthi, and Ng]{mildenhall2020nerf}
Ben Mildenhall, Pratul~P Srinivasan, Matthew Tancik, Jonathan~T Barron, Ravi Ramamoorthi, and Ren Ng.
\newblock Nerf: Representing scenes as neural radiance fields for view synthesis.
\newblock In \emph{ECCV}, 2020.

\bibitem[Mu et~al.(2024)Mu, Zuo, Guo, Wang, Lu, Wu, Xu, Dai, Yan, and Cheng]{mu2024gsd}
Yuxuan Mu, Xinxin Zuo, Chuan Guo, Yilin Wang, Juwei Lu, Xiaofeng Wu, Songcen Xu, Peng Dai, Youliang Yan, and Li Cheng.
\newblock {GSD}: View-guided gaussian splatting diffusion for {3D} reconstruction.
\newblock In \emph{ECCV}, 2024.

\bibitem[Park et~al.(2019)Park, Florence, Straub, Newcombe, and Lovegrove]{park2019deepsdf}
Jeong~Joon Park, Peter Florence, Julian Straub, Richard Newcombe, and Steven Lovegrove.
\newblock Deepsdf: Learning continuous signed distance functions for shape representation.
\newblock In \emph{CVPR}, 2019.

\bibitem[Peng et~al.(2025)Peng, Sobol, Tomizuka, Keutzer, Xu, and Litany]{peng2025lesson}
Chensheng Peng, Ido Sobol, Masayoshi Tomizuka, Kurt Keutzer, Chenfeng Xu, and Or Litany.
\newblock A lesson in splats: Teacher-guided diffusion for 3d gaussian splats generation with 2d supervision.
\newblock In \emph{ICCV}, 2025.

\bibitem[Poole et~al.(2023)Poole, Jain, Barron, and Mildenhall]{poole2022dreamfusion}
Ben Poole, Ajay Jain, Jonathan~T Barron, and Ben Mildenhall.
\newblock {DreamFusion}: Text-to-3d using 2d diffusion.
\newblock In \emph{ICLR}, 2023.

\bibitem[Radford et~al.(2021)Radford, Kim, Hallacy, Ramesh, Goh, Agarwal, Sastry, Askell, Mishkin, Clark, Krueger, and Sutskever]{radford2021learning}
Alec Radford, Jong~Wook Kim, Chris Hallacy, Aditya Ramesh, Gabriel Goh, Sandhini Agarwal, Girish Sastry, Amanda Askell, Pamela Mishkin, Jack Clark, Gretchen Krueger, and Ilya Sutskever.
\newblock Learning transferable visual models from natural language supervision.
\newblock In \emph{ICML}, 2021.

\bibitem[Reizenstein et~al.(2021)Reizenstein, Shapovalov, Henzler, Sbordone, Labatut, and Novotny]{reizenstein21co3d}
Jeremy Reizenstein, Roman Shapovalov, Philipp Henzler, Luca Sbordone, Patrick Labatut, and David Novotny.
\newblock Common objects in {3D}: Large-scale learning and evaluation of real-life {3D} category reconstruction.
\newblock In \emph{{ICCV}}, 2021.

\bibitem[Rombach et~al.(2022)Rombach, Blattmann, Lorenz, Esser, and Ommer]{rombach2022stablediffusion}
Robin Rombach, Andreas Blattmann, Dominik Lorenz, Patrick Esser, and Björn Ommer.
\newblock High-resolution image synthesis with latent diffusion models.
\newblock In \emph{{CVPR}}, 2022.

\bibitem[Ronneberger et~al.(2015)Ronneberger, Fischer, and Brox]{ronneberger2015u}
Olaf Ronneberger, Philipp Fischer, and Thomas Brox.
\newblock U-net: Convolutional networks for biomedical image segmentation.
\newblock In \emph{MICCAI}, 2015.

\bibitem[Sargent et~al.(2024)Sargent, Li, Shah, Herrmann, Yu, Zhang, Chan, Lagun, Fei-Fei, Sun, and Wu]{sargent2024zeronvs}
Kyle Sargent, Zizhang Li, Tanmay Shah, Charles Herrmann, Hong-Xing Yu, Yunzhi Zhang, Eric~Ryan Chan, Dmitry Lagun, Li Fei-Fei, Deqing Sun, and Jiajun Wu.
\newblock {ZeroNVS}: Zero-shot 360-degree view synthesis from a single image.
\newblock In \emph{CVPR}, 2024.

\bibitem[Sch\"{o}nberger and Frahm(2016)]{schoenberger2016sfm}
Johannes~Lutz Sch\"{o}nberger and Jan-Michael Frahm.
\newblock Structure-from-motion revisited.
\newblock In \emph{CVPR}, 2016.

\bibitem[Schwarz et~al.(2025)Schwarz, Mueller, and Kontschieder]{schwarz2025generative}
Katja Schwarz, Norman Mueller, and Peter Kontschieder.
\newblock Generative gaussian splatting: Generating 3d scenes with video diffusion priors.
\newblock In \emph{ICCV}, 2025.

\bibitem[Shen et~al.(2025)Shen, Wu, Yi, Zhou, Zhang, Yan, and Wang]{shen2025gamba}
Qiuhong Shen, Zike Wu, Xuanyu Yi, Pan Zhou, Hanwang Zhang, Shuicheng Yan, and Xinchao Wang.
\newblock Gamba: Marry gaussian splatting with mamba for single-view 3d reconstruction.
\newblock \emph{TPAMI}, 2025.

\bibitem[Song et~al.(2017)Song, Yu, Zeng, Chang, Savva, and Funkhouser]{song2017semantic}
Shuran Song, Fisher Yu, Andy Zeng, Angel~X Chang, Manolis Savva, and Thomas Funkhouser.
\newblock Semantic scene completion from a single depth image.
\newblock In \emph{{CVPR}}, 2017.

\bibitem[Szymanowicz et~al.(2024{\natexlab{a}})Szymanowicz, Insafutdinov, Zheng, Campbell, Henriques, Rupprecht, and Vedaldi]{szymanowicz2024flash3d}
Stanislaw Szymanowicz, Eldar Insafutdinov, Chuanxia Zheng, Dylan Campbell, Jo{\~a}o~F Henriques, Christian Rupprecht, and Andrea Vedaldi.
\newblock {Flash3D}: Feed-forward generalisable {3D} scene reconstruction from a single image.
\newblock \emph{arXiv}, 2024{\natexlab{a}}.

\bibitem[Szymanowicz et~al.(2024{\natexlab{b}})Szymanowicz, Rupprecht, and Vedaldi]{szymanowicz2024splatter}
Stanislaw Szymanowicz, Chrisitian Rupprecht, and Andrea Vedaldi.
\newblock Splatter image: Ultra-fast single-view {3D} reconstruction.
\newblock In \emph{{CVPR}}, 2024{\natexlab{b}}.

\bibitem[Szymanowicz et~al.(2025)Szymanowicz, Zhang, Srinivasan, Gao, Brussee, Holynski, Martin-Brualla, Barron, and Henzler]{szymanowicz2025bolt3d}
Stanislaw Szymanowicz, Jason~Y Zhang, Pratul Srinivasan, Ruiqi Gao, Arthur Brussee, Aleksander Holynski, Ricardo Martin-Brualla, Jonathan~T Barron, and Philipp Henzler.
\newblock Bolt3d: Generating 3d scenes in seconds.
\newblock In \emph{{ICCV}}, 2025.

\bibitem[Tang et~al.(2023)Tang, Ren, Zhou, Liu, and Zeng]{tang2023dreamgaussian}
Jiaxiang Tang, Jiawei Ren, Hang Zhou, Ziwei Liu, and Gang Zeng.
\newblock {DreamGaussian}: Generative gaussian splatting for efficient {3D} content creation.
\newblock In \emph{ICLR}, 2023.

\bibitem[Tang et~al.(2024)Tang, Chen, Chen, Wang, Zeng, and Liu]{tang2024lgm}
Jiaxiang Tang, Zhaoxi Chen, Xiaokang Chen, Tengfei Wang, Gang Zeng, and Ziwei Liu.
\newblock {LGM}: Large multi-view gaussian model for high-resolution 3d content creation.
\newblock In \emph{ECCV}, 2024.

\bibitem[Tewari et~al.(2023)Tewari, Yin, Cazenavette, Rezchikov, Tenenbaum, Durand, Freeman, and Sitzmann]{tewari2023diffusion}
Ayush Tewari, Tianwei Yin, George Cazenavette, Semon Rezchikov, Josh Tenenbaum, Fr{\'e}do Durand, Bill Freeman, and Vincent Sitzmann.
\newblock Diffusion with forward models: Solving stochastic inverse problems without direct supervision.
\newblock \emph{NeurIPS}, 2023.

\bibitem[Wang et~al.(2025)Wang, Chen, Karaev, Vedaldi, Rupprecht, and Novotny]{wang2025vggt}
Jianyuan Wang, Minghao Chen, Nikita Karaev, Andrea Vedaldi, Christian Rupprecht, and David Novotny.
\newblock {VGGT}: Visual geometry grounded transformer.
\newblock In \emph{CVPR}, 2025.

\bibitem[Wang et~al.(2024)Wang, Leroy, Cabon, Chidlovskii, and Revaud]{wang2024dust3r}
Shuzhe Wang, Vincent Leroy, Yohann Cabon, Boris Chidlovskii, and Jerome Revaud.
\newblock {DUSt3R}: Geometric 3d vision made easy.
\newblock In \emph{CVPR}, 2024.

\bibitem[Wang et~al.(2017)Wang, Huang, You, Yang, and Neumann]{wang2017shape}
Weiyue Wang, Qiangui Huang, Suya You, Chao Yang, and Ulrich Neumann.
\newblock Shape inpainting using {3D} generative adversarial network and recurrent convolutional networks.
\newblock In \emph{ICCV}, 2017.

\bibitem[Wang et~al.(2004)Wang, Bovik, Sheikh, and Simoncelli]{wang2004image}
Zhou Wang, Alan~C Bovik, Hamid~R Sheikh, and Eero~P Simoncelli.
\newblock Image quality assessment: from error visibility to structural similarity.
\newblock \emph{IEEE TIP}, 2004.

\bibitem[Wewer et~al.(2024)Wewer, Raj, Ilg, Schiele, and Lenssen]{wewer24latentsplat}
Christopher Wewer, Kevin Raj, Eddy Ilg, Bernt Schiele, and Jan~Eric Lenssen.
\newblock {latentSplat}: Autoencoding variational gaussians for fast generalizable {3D} reconstruction.
\newblock In \emph{ECCV}, 2024.

\bibitem[Wu et~al.(2015)Wu, Song, Khosla, Yu, Zhang, Tang, and Xiao]{wu20153d}
Zhirong Wu, Shuran Song, Aditya Khosla, Fisher Yu, Linguang Zhang, Xiaoou Tang, and Jianxiong Xiao.
\newblock {3D ShapeNets}: A deep representation for volumetric shapes.
\newblock In \emph{{CVPR}}, 2015.

\bibitem[Xiang et~al.(2025)Xiang, Li, Long, H{\"a}ne, Guo, Delp, Adeli, and Fei-Fei]{xiang2025repurposing}
Tiange Xiang, Kai Li, Chengjiang Long, Christian H{\"a}ne, Peihong Guo, Scott Delp, Ehsan Adeli, and Li Fei-Fei.
\newblock Repurposing 2d diffusion models with gaussian atlas for 3d generation.
\newblock In \emph{{ICCV}}, 2025.

\bibitem[Xu et~al.(2024{\natexlab{a}})Xu, Yuan, Mardani, Liu, Song, Wang, and Vahdat]{xu2024agg}
Dejia Xu, Ye Yuan, Morteza Mardani, Sifei Liu, Jiaming Song, Zhangyang Wang, and Arash Vahdat.
\newblock {AGG}: Amortized generative {3D} gaussians for single image to {3D}.
\newblock \emph{TMLR}, 2024{\natexlab{a}}.

\bibitem[Xu et~al.(2025)Xu, Peng, Wang, Blum, Barath, Geiger, and Pollefeys]{xu2024depthsplat}
Haofei Xu, Songyou Peng, Fangjinhua Wang, Hermann Blum, Daniel Barath, Andreas Geiger, and Marc Pollefeys.
\newblock {DepthSplat}: Connecting gaussian splatting and depth.
\newblock In \emph{CVPR}, 2025.

\bibitem[Xu et~al.(2024{\natexlab{b}})Xu, Shi, Yifan, Chen, Yang, Peng, Shen, and Wetzstein]{xu2024grm}
Yinghao Xu, Zifan Shi, Wang Yifan, Hansheng Chen, Ceyuan Yang, Sida Peng, Yujun Shen, and Gordon Wetzstein.
\newblock {GRM}: Large gaussian reconstruction model for efficient 3d reconstruction and generation.
\newblock In \emph{ECCV}, 2024{\natexlab{b}}.

\bibitem[Xu et~al.(2024{\natexlab{c}})Xu, Tan, Luan, Bi, Wang, Li, Shi, Sunkavalli, Wetzstein, Xu, and Zhang]{xu2023dmv3d}
Yinghao Xu, Hao Tan, Fujun Luan, Sai Bi, Peng Wang, Jiahao Li, Zifan Shi, Kalyan Sunkavalli, Gordon Wetzstein, Zexiang Xu, and Kai Zhang.
\newblock {DMV3D}: Denoising multi-view diffusion using 3d large reconstruction model.
\newblock In \emph{ICLR}, 2024{\natexlab{c}}.

\bibitem[Yi et~al.(2024)Yi, Fang, Wang, Wu, Xie, Zhang, Liu, Tian, and Wang]{yi2024gaussiandreamer}
Taoran Yi, Jiemin Fang, Junjie Wang, Guanjun Wu, Lingxi Xie, Xiaopeng Zhang, Wenyu Liu, Qi Tian, and Xinggang Wang.
\newblock Gaussiandreamer: Fast generation from text to {3D} gaussians by bridging {2D} and {3D} diffusion models.
\newblock In \emph{CVPR}, 2024.

\bibitem[Yu et~al.(2021)Yu, Ye, Tancik, and Kanazawa]{yu2021pixelnerf}
Alex Yu, Vickie Ye, Matthew Tancik, and Angjoo Kanazawa.
\newblock {pixelNeRF}: Neural radiance fields from one or few images.
\newblock In \emph{{CVPR}}, 2021.

\bibitem[Zhang et~al.(2024{\natexlab{a}})Zhang, Song, Wei, Chen, Lu, and Tang]{zhang2024geolrm}
Chubin Zhang, Hongliang Song, Yi Wei, Yu Chen, Jiwen Lu, and Yansong Tang.
\newblock {GeoLRM}: Geometry-aware large reconstruction model for high-quality {3D} gaussian generation.
\newblock In \emph{NeurIPS}, 2024{\natexlab{a}}.

\bibitem[Zhang et~al.(2024{\natexlab{b}})Zhang, Bi, Tan, Xiangli, Zhao, Sunkavalli, and Xu]{zhang2024gs}
Kai Zhang, Sai Bi, Hao Tan, Yuanbo Xiangli, Nanxuan Zhao, Kalyan Sunkavalli, and Zexiang Xu.
\newblock {GS-LRM}: Large reconstruction model for 3d gaussian splatting.
\newblock In \emph{ECCV}, 2024{\natexlab{b}}.

\bibitem[Zhang et~al.(2018)Zhang, Isola, Efros, Shechtman, and Wang]{zhang2018perceptual}
Richard Zhang, Phillip Isola, Alexei~A Efros, Eli Shechtman, and Oliver Wang.
\newblock The unreasonable effectiveness of deep features as a perceptual metric.
\newblock In \emph{{CVPR}}, 2018.

\bibitem[Zhang et~al.(2025)Zhang, Li, Fei, Liu, and Duan]{zhang2025scene}
Shengjun Zhang, Jinzhao Li, Xin Fei, Hao Liu, and Yueqi Duan.
\newblock Scene splatter: Momentum 3d scene generation from single image with video diffusion model.
\newblock In \emph{CVPR}, 2025.

\bibitem[Zhou et~al.(2018)Zhou, Tucker, Flynn, Fyffe, and Snavely]{zhou2018stereo}
Tinghui Zhou, Richard Tucker, John Flynn, Graham Fyffe, and Noah Snavely.
\newblock Stereo magnification: Learning view synthesis using multiplane images.
\newblock \emph{SIGGRAPH}, 2018.

\bibitem[Zhou and Tulsiani(2023)]{zhou2023sparsefusion}
Zhizhuo Zhou and Shubham Tulsiani.
\newblock {SparseFusion}: Distilling view-conditioned diffusion for {3D} reconstruction.
\newblock In \emph{{CVPR}}, 2023.

\end{thebibliography}
}

\newpage

\appendix

\counterwithin{figure}{section}
\counterwithin{table}{section}
\renewcommand\thefigure{\thesection\arabic{figure}}
\renewcommand\thetable{\thesection\arabic{table}}

\section{Methods Details}

\subsection{Gaussian Splatting Representation}
\label{supp:splatterimage_representation}

We represent scenes as a set of Gaussian splats~\cite{kerbl3Dgaussians}, and follow Splatter Image~\cite{szymanowicz2024splatter} to represent Gaussians inside a frustum. In this section, we first introduce the definitions of Gaussian splatting, then describe how to represent Gaussians within an image frustum, and how a neural network is used to predict them.

\paragraph{Gaussian Splatting}
Following~\cite{kerbl3Dgaussians}, a 3D scene is parameterized as a set of Gaussian splats, each of which is associated with an opacity $\opacity$, 3D position $\mean$, scale and rotation parameters of the covariance matrix, and color $\colorsh$ modeled with spherical harmonics. Through differentiable rendering, all Gaussians can be projected onto image planes to form RGB images. Typically, the parameters of the Gaussian splats are optimized to match the renderings with a set of posed RGB training images, in order to reconstruct a single scene at a time.

\paragraph{Splatter Image}
The ``Splatter Image''~\cite{szymanowicz2024splatter} formulation predicts a Gaussian for each pixel ray of a single $H \times W$ input RGB image $I$, representing both observed and unobserved scene content inside the image frustum.  
This results in a parameter matrix of shape $H \times W \times N$, where $N$ is the number of parameters per Gaussian splat.  
Specifically, the Gaussian mean is parameterized by the pixel depth and a 3D offset.  
Each Gaussian thus contains a total of $N = 15$ scalar values: 1 (opacity), 1 (depth), 3 (offset), 3 (scale), 4 (rotation quaternion), and 3 (color).  
We use only the DC coefficients for spherical harmonics.
To further encourage the network to model occluded surfaces, follow-up work~\cite{szymanowicz2024flash3d} extends the parameter matrix to $H \times W \times MN$, where $M$ denotes the number of Gaussian splats predicted per pixel, i.e., the number of Gaussian layers.

\paragraph{Training a Network to Predict Splatter Image}
Following~\cite{szymanowicz2024splatter}, a neural network can be trained to directly predict the Gaussian splat parameters of a Splatter Image via a feed-forward process. Given a single input image, the network (e.g., a U-Net) outputs the complete parameter matrix corresponding to the Gaussian splats, while preserving the spatial structure of the image.
However, this network is conventionally regression-based and can only model a single possibility~\cite{szymanowicz2024splatter,charatan2024pixelsplat,szymanowicz2024flash3d}. In contrast, we extend this formulation to a distribution using a diffusion model, enabling the modeling of multiple plausible outputs.

\subsection{Diffusion Pipelines}

\begin{figure*}
    \centering
    \includegraphics[width=0.9\textwidth]{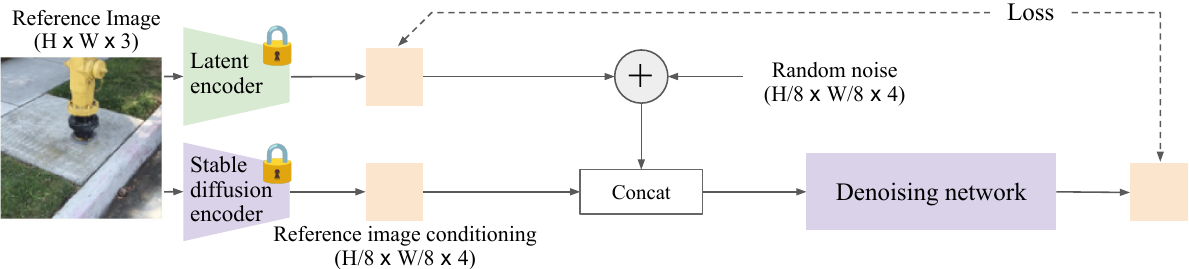}
    \caption[Training a denoising diffusion model over the learned latent space.]{\textbf{Training a denoising diffusion model over the learned latent space.}  
The network learns to convert a corrupted noisy latent code back to a ground-truth latent code representing a Splatter Image, by conditioning on a single input image.
    }

    \label{fig:training_step2}
\end{figure*}

\begin{figure*}
    \includegraphics[width=\textwidth]{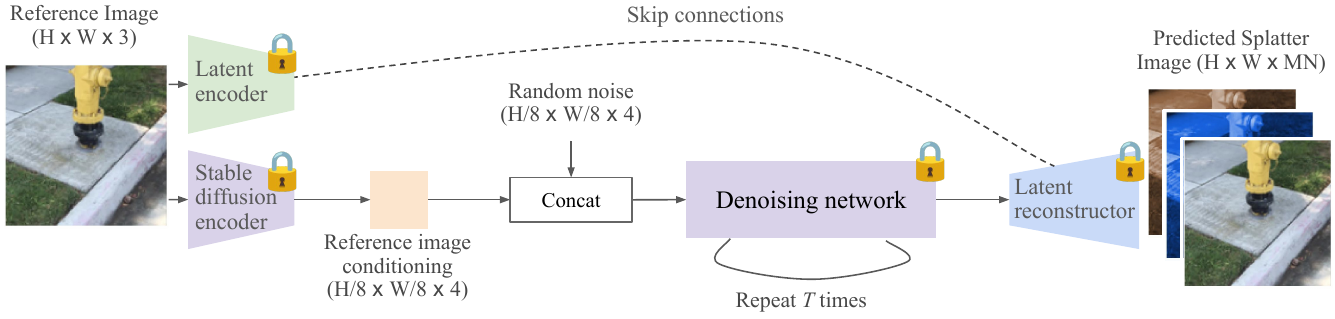}
    \vspace{-18pt}
    \caption[Diffusion Inference Pipeline.]{\textbf{Diffusion Inference Pipeline.}  
We first compute input image features using the Stable Diffusion encoder. A random latent code and the input image features are concatenated and passed through \( R \) steps of the denoising diffusion process. The denoised latent code, together with skip connections from the encoder, is then passed through the reconstructor to produce a Splatter Image representation. This representation is subsequently backprojected to create 3D Gaussian splats.
        }
    \vspace{-6pt}
    \label{fig:inference}
\end{figure*}

To better illustrate our diffusion architecture, we present figures for diffusion training over the learned latent space in Figure~\ref{fig:training_step2}, and the diffusion inference procedure in Figure~\ref{fig:inference}.

\subsection{Alternative Approaches for Learning a Latent Space for Splatter Images}
\label{section:supp:alternative_latent_space_learning}

Our Variational AutoReconstructor learns a latent space of Splatter Images in a simple and effective way.  
We further describe alternative approaches we explored in the early stages of the project that did not yield satisfactory results. 

\paragraph{Reconstructor-only without the Encoder} Since we only need latents for diffusion training, we try removing the encoder. We assign each image an optimizable latent code and optimize both the latent code and the reconstructor weights to generate Splatter Images, similar to the autodecoder approach for SDF representations~\cite{park2019deepsdf}. However, we observe that this leads to low-quality reconstructions because the optimization gets stuck in the early stages and fails to capture the geometry correctly. We assume the encoder is important for initializing the latents or capturing correlations that are crucial to the input images.

\paragraph{Pre-optimizing 3D Gaussian splats and projecting into Splatter Images}  This approach optimizes 3D Gaussian splats for each scene using all available images, then projects the Gaussians onto each image plane to form Splatter Images. However, rasterization errors occur when projecting Gaussians to achieve pixel alignment, and the 3D Gaussians are not distributed uniformly across the images.  
The resulting Splatter Images are extremely sparse, with 80\% of the pixels lacking a corresponding Gaussian.  
We find it difficult to train a VAE to encode each modalities of such sparse splatter images and reconstruct them with negligible error, which breaks the geometric and texture consistency.

\section{Implementation Details}

\subsection{Network architecture}
We implement the AutoReconstructor architecture following the VAE from Stable Diffusion~\cite{rombach2022stablediffusion}. The encoder remains unchanged and predicts a distribution over the latent representation at \( \tfrac{1}{8} \) the resolution of the input images, with \( 4 + 4 \) channels (4 for \( \code_\mu \) and 4 for \( \code_\sigma \)), from which a 4-channel latent can be sampled.
We extend the decoder into a reconstructor by modifying its output layers: instead of producing 3 channels for RGB images, it now outputs \( MN \) channels to represent Gaussians for each pixel. We predict \( M = 2 \) Gaussians for each pixel, and set \( N = 15 \) as described in Section~\ref{supp:splatterimage_representation}.
We train the AutoReconstructor by fine-tuning it based on the original VAE parameters.
We follow the same denoiser architecture as Stable Diffusion~\cite{rombach2022stablediffusion} and train it from scratch.

\subsection{Data Preprocessing}
For a fair comparison with prior work~\cite{chan2023genvs, szymanowicz2024splatter, tewari2023diffusion}, all images in the datasets are cropped around the principal point of the camera (using the largest possible crop) and rescaled to \(128 \times 128\) pixels before training.

\paragraph{Scaling of scenes}
We follow the approach outlined in the Splatter Image~\cite{szymanowicz2024splatter} supplementary material (Section B.2) to address varying scene scales. In practice, this involves rescaling the predicted Gaussian depths (at both training and test time) to lie between \( z_\text{near} \) and \( z_\text{far} \), where \( z_\text{near} \) and \( z_\text{far} \) are computed using the known depth to the object in the scene.

\subsection{Training Details}

\paragraph{Encoder-Reconstructor}
We first train the AutoReconstructor. We use batch size 32 and Adam optimizer.
We train for 6000 epochs with L2 and SSIM losses ($\lambda_1=0.8, \lambda_2=0.2$) and learning rate of 5e-5.
Fine-tuning is done with another 8000 epochs with L2, SSIM and LPIPS losses ($\lambda_1=0.8, \lambda_2=0.2, \lambda_3=0.01$) and learning rate of 5e-6. Using LPIPS loss only during finetuning is a common practice~\cite{szymanowicz2024splatter} to prevent the LPIPS loss from disrupting geometric structure learning in the early stages of training.
We apply the same random color augmentation to both the reference and target views to increase the diversity of colors that the latent codes can represent.

\paragraph{Skip connection}
When using the skip connection, we observed that the model can ``cheat'' by retaining most of the information about the scene in the skip connection features instead of latent codes. 
To limit this effect, during training we randomly zero out all the values of the skip features, encouraging the network to encapsulate scene information in the latent codes.

\paragraph{Denoising Diffusion Model}
We train with Adam Optimizer and batch size 32.
We train for $1,000$K steps and a probability of 20\% for unconditional generation (classifier-free guidance).
We finetune for $100$K steps with a probability of 50\% for randomly masking the input images.
Learning rate is set to 1e-4 with exponential learning rate scheduler, similar to Marigold~\cite{ke2023repurposing}.
We also sample latent codes from the encoder network as ground truth latent codes during diffusion trainings.

\section{Experiments Details}

\subsection{Evaluation Protocol}
After training on the CO3D or RealEstate10K datasets, we evaluate on the unseen test sets. For each sequence, we use the reference view to predict Gaussian splats and render them from both the reference viewpoint and several novel target viewpoints. We then compare the rendered images with the corresponding ground-truth RGB images captured from the same viewpoints.
We design two settings for comprehensive evaluations: \textit{Full Images} and \textit{Objects Only}. 
\textit{Full Images}, computes the metrics on all pixels of reference views.
The \textit{Objects Only} evaluation computes the metrics only on object pixels, following the protocol of Splatter Image~\cite{szymanowicz2024splatter}, and is reported here for a fair comparison with prior work.
Similar to our method, DFM~\cite{tewari2023diffusion} and Splatter Image (Full Images)\(\star\) always predict background pixels. For the \textit{Objects Only} evaluation, we use the ground truth object mask to set the background pixels of both the predicted and ground-truth images to black before computing the metrics.

\subsection{Baselines}

\paragraph{Open-source methods}  
We present the open-source status of the literature methods in Table~\ref{sup:table:baselines_open_source}, from which we select our main competitors in the experiments for evaluation.

\begin{table}
\centering
\resizebox{0.9\columnwidth}{!}{
\begin{tabular}{@{}lcc@{}}
\toprule
\textbf{Methods} & \textbf{Open-source Code} & \textbf{Baselines Used} \\ \midrule
DFM~\cite{tewari2023diffusion}              & \checkmark                  & \checkmark                \\
SplatterImage~\cite{szymanowicz2024splatter}    & \checkmark                  & \checkmark                \\
pixelNeRF~\cite{yu2021pixelnerf}        & \checkmark                  & \checkmark                \\
SparseFusion~\cite{zhou2023sparsefusion}     & \checkmark                  & \checkmark                \\
ZeroNVS~\cite{sargent2024zeronvs}          & \checkmark                  & \checkmark                \\
LGM~\cite{tang2024lgm}              & \checkmark                  & \checkmark                \\
Diffsplat~\cite{lin2025diffsplat} *       & \checkmark                  & $\times$                 \\
SampleSplat~\cite{henderson2024sampling}      & $\times$                 &    $\times$                \\
Bolt3D~\cite{szymanowicz2025bolt3d}           & $\times$                 &    $\times$                \\
GRM~\cite{xu2024grm}              & $\times$                 &           $\times$         \\
GS-LRM~\cite{zhang2024gs}  $\dagger$         & $\times$                 &   $\times$                 \\
Zero-1-to-G~\cite{meng2025zero}      & $\times$                 &       $\times$             \\
Xiang et al.~\cite{xiang2025repurposing}      & $\times$                 &   $\times$                 \\ \bottomrule
\end{tabular}
}
\caption{\textbf{Availability of an open-source implementation for different related methods.} We choose our baselines from the closest competitors that provide open-source code and support scene-level reconstruction. *: Diffsplat is an object-centric method; $\dagger$: Only an incomplete unofficial version is available. Code availability checked on August 15, 2025.}
\label{sup:table:baselines_open_source}
\end{table}

\paragraph{Chosen Baselines}
We introduce in detail the baselines to which we compare below:

(1) \textit{Splatter Image}~\cite{szymanowicz2024splatter} is the closest model to ours, employing a regression-based formulation. We use the publicly available code and models for our experiments. This model was trained only on foreground objects by masking out the background.

(2) \textit{Splatter Image (Full Images)}\(\star\): To obtain a baseline more comparable to our method, which models the full scene, we introduce a modified and retrained version of the original Splatter Image~\cite{szymanowicz2024splatter} that predicts both objects and background by training on full images. The \(\star\) indicates that this model was trained by us. As a regression-based model, it produces uni-modal outputs and does not support sampling.

(3) \textit{PixelNeRF}~\cite{yu2021pixelnerf} is a NeRF-based regression method for novel view prediction conditioned on one or more images.

(4) \textit{DFM}~\cite{tewari2023diffusion} is a diffusion-based method built on NeRF that can generate high-quality renderings. However, the average scene takes approximately 10 minutes to reconstruct and render. We use the official codebase to produce outputs for evaluation. It includes models trained on the Hydrants and RealEstate10K datasets.

(5) \textit{ZeroNVS}~\cite{sargent2024zeronvs} employs diffusion and Score Distillation Sampling with anchoring to \textit{optimize} a NeRF, but it faces 3D consistency issues such as Multi-Face Janus artifacts~\cite{poole2022dreamfusion}. 

(6) \textit{SparseFusion}~\cite{zhou2023sparsefusion} outputs novel views of 3D scenes using a generative model to provide score distillation for each view, requiring intensive computation during optimization. We compare our method with ZeroNVS and SparseFusion on the RealEstate10K dataset.

(7) \textit{LGM}~\cite{tang2024lgm}: A large-scale object reconstruction model based on Gaussian splats. We demonstrate our object reconstruction performance, while additionally reconstructing background regions.

\subsection{Ablation Studies Analysis}

The results of the ablation study are shown in Table~\ref{tab:ablations}, and we discuss them in detail below:

\paragraph{Skip connections} 
\textit{No skip connections}: We ablate the variant that does not use skip connections, which also qualitatively shows degradation in Fig.~\ref{fig:skip-connection}.  
\textit{Skips from first 2 layers}: We further ablate by using only two skip connections from the first and second layers of the encoder, which may reduce the scene information learned in the latent codes. 

\paragraph{Number of Gaussians per Pixel} 
\textit{One Gaussian splat per pixel}: Our method trained to predict only a single Gaussian splat per pixel, which lacks sufficient Gaussians to represent scene details. 

\paragraph{Variational Sampling} 
\textit{No variational sampling}: Disables variational sampling and directly uses the 4-dimensional code regressed by the encoder. We observe that training gets stuck and does not improve in the early stages. 

\paragraph{Training Losses} 
We ablate training with \textit{No SSIM} or \textit{No LPIPS}, which results in critical performance drops. 

\paragraph{Data Augmentation} 
\textit{No Color Augmentation}: Does not apply color augmentation to the input and target images during training.

\subsection{3D Generative Experiments}
We present additional details of the 3D generative experiments in Figure~\ref{fig:results_3d_generation}. We train the diffusion model using random masking, while keeping the Variational AutoReconstructor model fixed. The masked input image is passed through the diffusion model to generate a latent code, which is then reconstructed into a Splatter Image using the pretrained AutoReconstructor. To focus on the generative capacity of the diffusion model, we disable skip connections during these experiments. Our goal is to demonstrate the model’s ability to produce diverse outputs from its latent space. To further enhance high-frequency details, techniques such as finetuning the AutoReconstructor with random masking could be explored, which we leave as future work.

\subsection{Experiments Notes}
\paragraph{Cited Baseline Scores}
In the quantitative results, we cite the reported scores from the papers of the baselines and ensure that the settings are consistent for a fair comparison.  
The main baseline scores are reported in Splatter Image~\cite{szymanowicz2024splatter} and DFM~\cite{tewari2023diffusion}.  
Some combinations of training data and methods are not publicly available for citation or evaluation, \eg, the performance of the DFM model trained on TeddyBears is not reported, and the corresponding model is not publicly available. Thus, we leave the corresponding entries blank in the table.

\section{Additional Experiments and Analysis}

\subsection{Objects-level Evaluations}

Results for \emph{Objects Only} in Table~\ref{tab:masked_object_results} provide a broader comparison to methods focused solely on objects, where our method outperforms almost all of them.  
Splatter Image slightly outperforms on TeddyBear LPIPS and KID metrics, partly due to its object-oriented design, while our approach focuses on the whole scene. We present a qualitative evaluation on TeddyBears in Figure~\ref{fig:results_teddybears}, where our method achieves more sensible reconstruction, especially in occluded areas.

\begin{figure}
    \centering
    \includegraphics[width=\columnwidth]{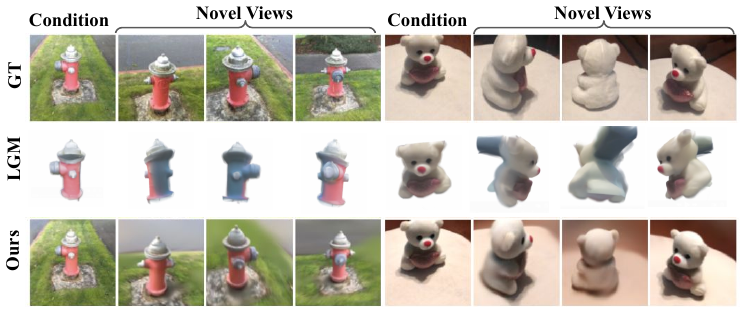}
    \caption[Comparison to LGM, a large-scale object reconstruction model using Gaussian splatting.]{ Comparison to LGM~\cite{tang2024lgm}, a large-scale object reconstruction model using Gaussian splatting, evaluated on the CO3D dataset. Our method outputs high-quality scene-level results with correct geometry and textures, even in occluded areas.
}
    \vspace{-8pt}
    \label{fig:lgm_compare}
\end{figure}

We further compare our method with representative large-scale object reconstruction approaches. Some methods~\cite{xu2024grm, zhang2024gs} do not have official code, pretrained models, or output results available for comparison, as also noted in Table~\ref{sup:table:baselines_open_source}. We qualitatively compare with LGM~\cite{tang2024lgm} in Figure~\ref{fig:results_teddybears} and Figure~\ref{fig:lgm_compare}, where our method demonstrates superior reconstruction quality, especially in preserving background context.

\begin{table*}
    \centering
    \resizebox{0.8\linewidth}{!}{
            \begin{tabular}{@{}lcccccccc@{}}
            \toprule
            \multirow{2}{*}{\textbf{Methods}}                                                        & \multicolumn{4}{c}{\textbf{Hydrants}}                                & \multicolumn{4}{c}{\textbf{Teddybears}}                              \\ \cmidrule(l){2-9} 
                                                                                                     & \textbf{PSNR ↑} & \textbf{LPIPS ↓} & \textbf{FID ↓} & \textbf{KID ↓} & \textbf{PSNR ↑} & \textbf{LPIPS ↓} & \textbf{FID ↓} & \textbf{KID ↓} \\ \midrule
            pi-GAN \cite{chanmonteiro2020piGAN} $\dagger$                           & -               & -                & 92.1           & 0.080          & -               & -                & 125.8          & 0.118          \\
            EG3D \cite{Chan2021} $\dagger$                                          & -               & -                & 229.5          & 0.253          & -               & -                & 236.1          & 0.239          \\
            GET3D \cite{gao2022get3d} $\dagger$                                     & -               & -                & 303.3          & 0.380          & -               & -                & 244.5          & 0.280          \\
            HoloDiffusion (no bootstrap) \cite{karnewar2023holodiffusion} $\dagger$ & -               & -                & 277.9          & 0.305          & -               & -                & 222.1          & 0.217          \\
            HoloDiffusion \cite{karnewar2023holodiffusion} $\dagger$                & -               & -                & 100.5          & 0.079          & -               & -                & 109.2          & 0.106          \\
            pixelNeRF \cite{yu2021pixelnerf} $\dagger$                              & 21.76           & 0.203            & -              & -              & 19.38           & 0.290            & -              & -              \\ \midrule
            Splatter Image (Objects Only) \cite{szymanowicz2024splatter} $\dagger$  & 21.80           & 0.150            & -              & -              & 19.44           & 0.231            & -              & -              \\ \midrule
            Splatter Image (Full Images) $\star$                                                     & 24.00           & 0.141            & 75.22          & 0.045          & {\ul 22.80}     & \textbf{0.172}   & \textbf{44.93} & 0.026          \\
            DFM~\cite{tewari2023diffusion}                                     & 23.37           & 0.133            & 60.1           & 0.029          & -               & -                & -              & -              \\
            Our AutoReconstructor $\star$                                                              & {\ul 24.18}     & 0.128            & 49.45          & 0.025          & \textbf{22.88}  & {\ul 0.176}      & {\ul 47.54}    & 0.026          \\
            Ours Diffusion single sample $\star$                                     & 23.87           & {\ul 0.127}      & {\ul 49.34}    & {\ul 0.024}    & 22.01           & 0.179            & 48.17          & 0.026          \\
            Ours Diffusion 20-best oracle $\star$                                    & \textbf{24.36}  & \textbf{0.124}   & \textbf{48.12} & \textbf{0.022} & 22.42           & {\ul 0.176}      & 47.62          & 0.027          \\ \bottomrule
            \end{tabular}
    }
    \caption[Quantitative results on masked objects from the CO3D dataset.]{\textbf{Quantitative results on masked objects from the CO3D dataset.}
    To compare against baselines which only predict novel views for the masked foreground object, as shown in the ``Splatter Image" row in Fig.~\ref{fig:results}, we masked the background areas and compare metrics. 
    $\dagger$: cited from respective papers or \cite{tewari2023diffusion}; $\star$: trained by us.
    }
    \label{tab:masked_object_results}
\end{table*}

\subsection{Additional Qualitative Results for Diverse Samplings}\label{sec:sup_diverse_sampling}

\begin{figure}
    \centering
    \includegraphics[width=0.95\columnwidth]{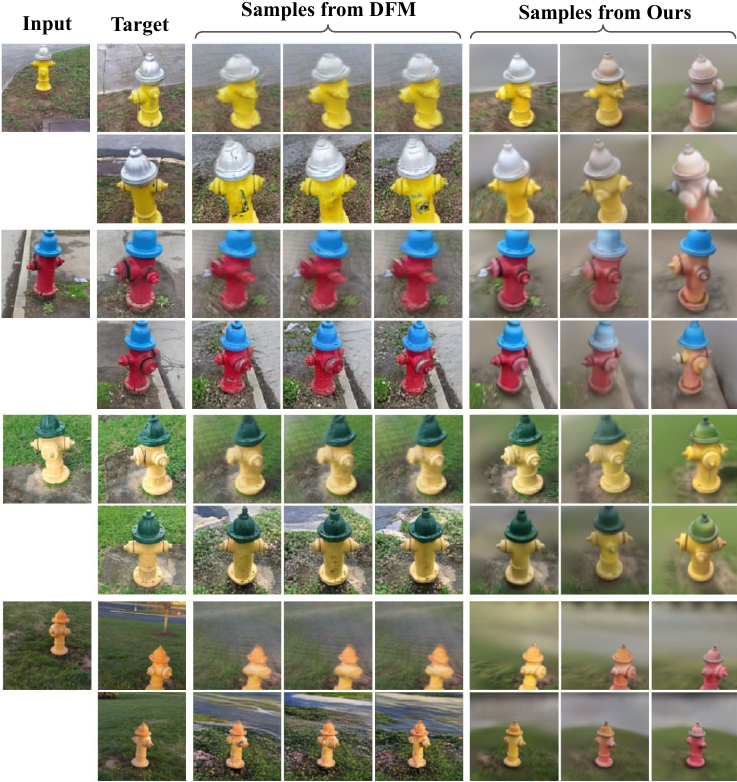}
    \caption[Diverse samples on Hydrants in the CO3D dataset.]{\textbf{Diverse samples from our diffusion model on Hydrants in the CO3D dataset}. 
We intentionally show three samples with increasing diversity from left to right by controlling the classifier-free guidance and skip connection weights. 
The samples transition from faithful reconstruction of the input image to diverse generations exhibiting variations in texture, shape, and style. 
In contrast, the baseline DFM~\cite{tewari2023diffusion} exhibits only small texture variations on object areas.
}
    \label{fig:diffusion_samples}
    \vspace{-8pt}
\end{figure}

We show more diverse sampling results on Hydrants in Figure~\ref{fig:diffusion_samples}.
For each input image, we show three generated reconstructions rendered from two different target views.
From left to right, the diversity \textit{increases} by manually \textit{increasing} the classifier-free guidance weight and \textit{decreasing} the skip connection weight.
Specifically, the three samples are generated with skip connection weights of (1.0, 0.5, 0.0) and classifier-free guidance weights of (0.2, 0.5, 0.5), respectively.  
This controllability allows tailoring the output to specific application needs, favoring high fidelity for reconstruction tasks, and greater diversity for applications like data augmentation or creative generation.

In the figures, we also compare with DFM~\cite{tewari2023diffusion}, which is a denoising diffusion model based on the NeRF representation, and show that our samples are much more diverse, especially in the occluded object areas. 
DFM shows small variations in the texture of objects, for example, some rust, while ours is able to add a valve structure and change the color. 
Furthermore, DFM denoises 2D images view by view, whereas ours outputs the Gaussian splats in a single diffusion process and is much faster.

\subsection{Additional Qualitative Results for Reconstruction}

\begin{figure*}
    \includegraphics[width=\textwidth]{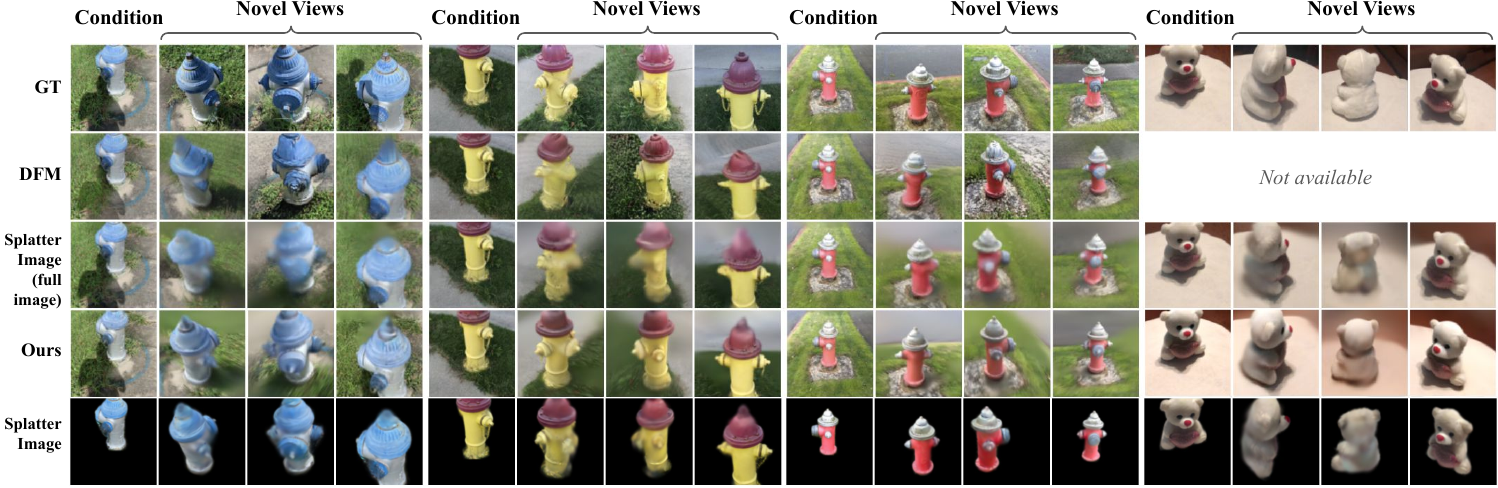}
    \caption[Qualitative results on the ``Hydrants'' and ``TeddyBears'' categories from the CO3D dataset.]{
        \textbf{Qualitative results} on the ``Hydrants'' and ``TeddyBears'' categories from the CO3D dataset. Our model produces sharper results with higher-quality details compared to the baselines, especially in occluded areas.
    }
    \label{fig:results}
\end{figure*}

\begin{figure*}
    \centering
    \includegraphics[width=1.0\textwidth]{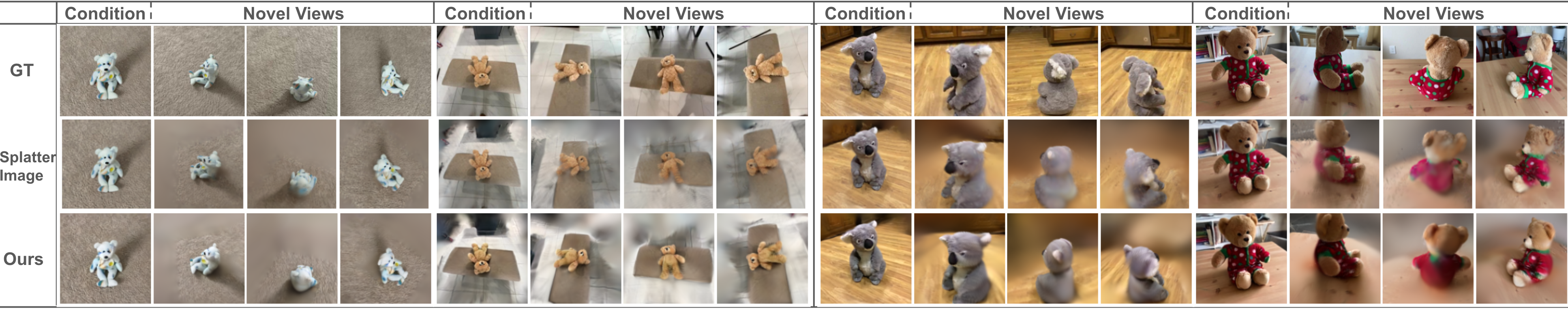}
    \includegraphics[width=1.0\textwidth]{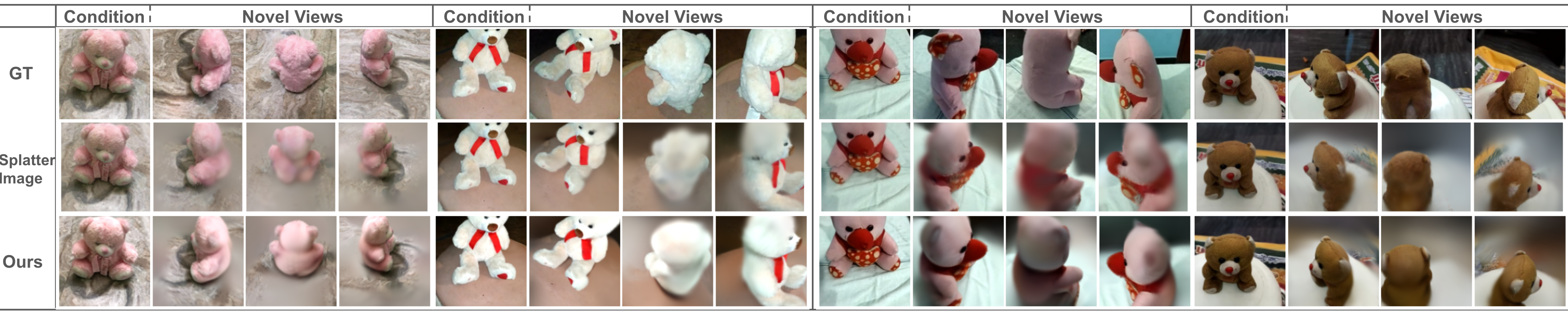}
    \includegraphics[width=1.0\textwidth]{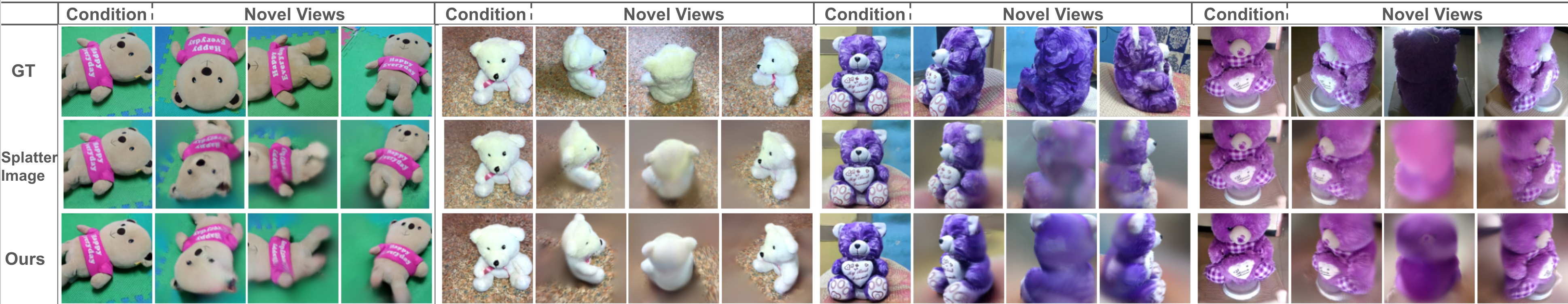}
    \caption{Additional qualitative results on ``TeddyBears''.}
    \label{fig:sup_teddies_results}
\end{figure*}

\begin{figure*}
    \centering
    \includegraphics[width=1.0\textwidth]{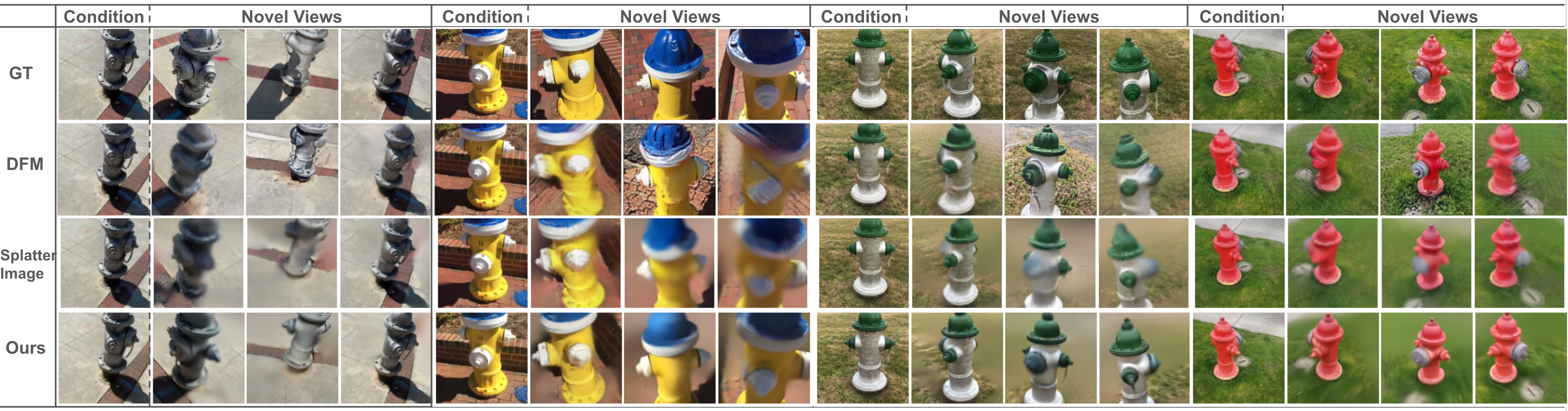}
    \includegraphics[width=1.0\textwidth]{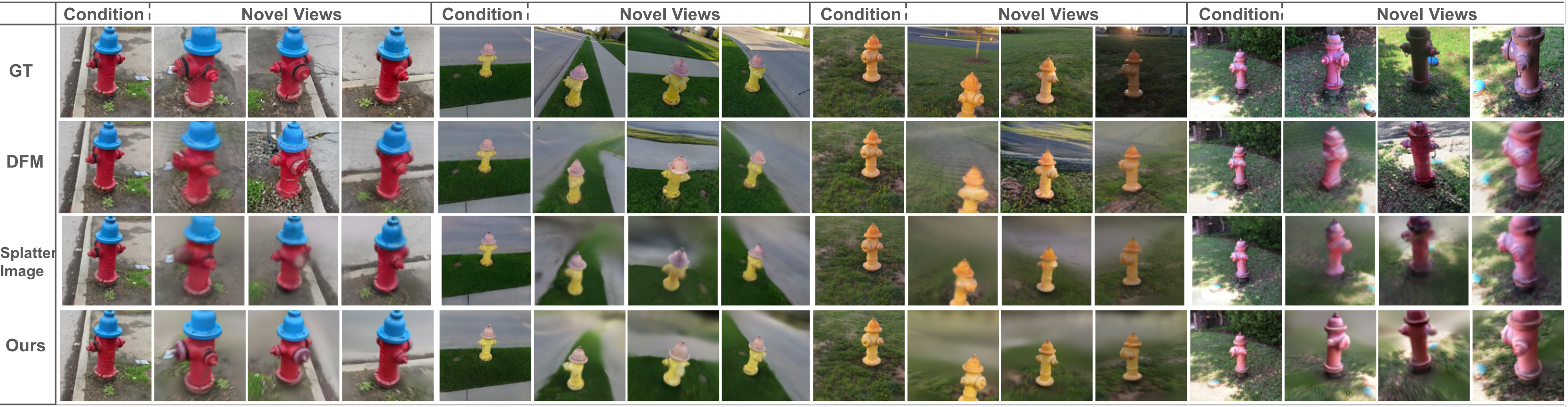}
    \includegraphics[width=1.0\textwidth]{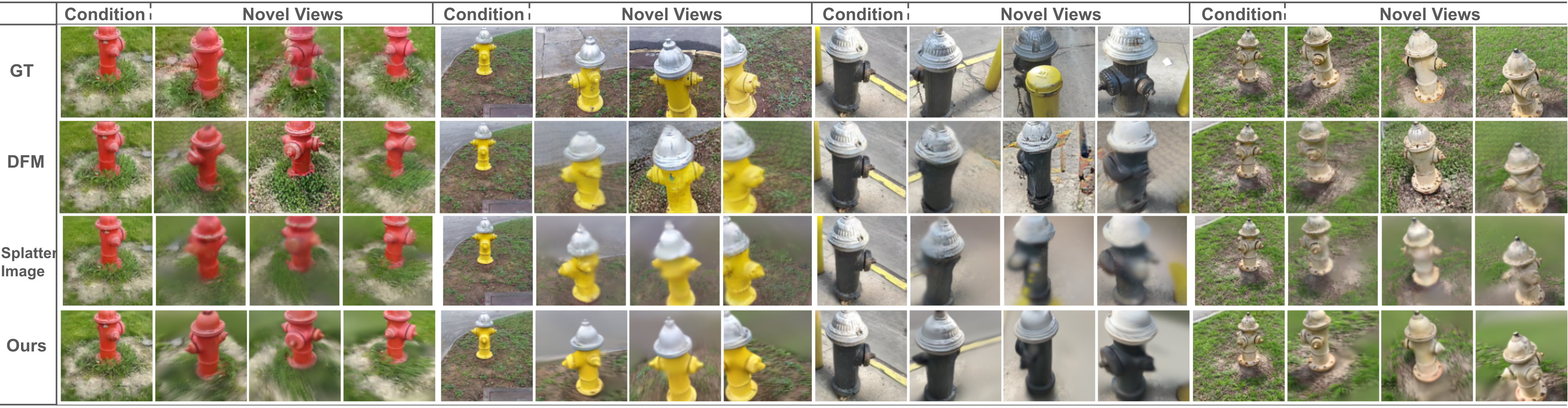}
    \caption{Additional qualitative results on ``Hydrants''.}
    \label{fig:sup_hydrants_results}
\end{figure*}

\begin{figure*}
    \centering
    \includegraphics[width=1.0\textwidth]{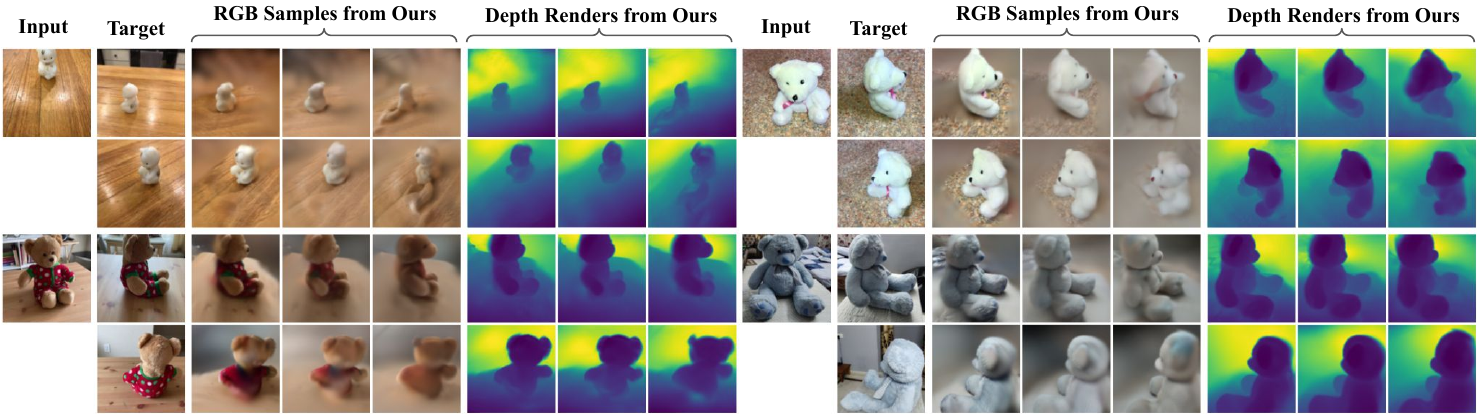}
    \caption{This image shows that our samples are diverse in both RGB and depth.
    }
    \label{fig:depth_diversity}
\end{figure*}

\begin{itemize}
    \item Figure~\ref{fig:results}, Figures~\ref{fig:sup_teddies_results} and~\ref{fig:sup_hydrants_results} show more reconstruction results on hydrants and teddybears.
    \item Figure~\ref{fig:depth_diversity} further shows renders of depth maps from our Gaussian splats, demonstrating that our samples are diverse in both geometry and appearance.
\end{itemize}

\subsection{Computation Analysis}
On each dataset, the AutoReconstructor takes 6 days and the diffusion model takes 2 days to train, both on \textbf{1} A100 GPU.  
Compared with the closest diffusion model DFM, which requires 7 days of training on \textbf{8} A100 GPUs, our method is significantly more efficient.  
During inference, as shown in Table~\ref{tab:whole_object_results}, generating a Splatter Image and rendering novel view images for one sequence takes only 3 seconds, while DFM requires 10 minutes.  
If using only the AutoReconstructor, the inference time can further decrease to 0.15 seconds.  
Overall, our pipeline is quite efficient in both training and inference, with strong potential to scale up to larger datasets and be used in online downstream applications.

\section{Limitations and Future Work} 

We make a step forward in training generative models on 3D scenes using Gaussian splatting representations. However, there are still limitations that suggest promising directions for future work. 
First, we model scenes using Splatter Images, which are largely constrained to the image frustum of a single view. Extending this representation to large-scale scenarios, either by exploring multiple Splatter Images or alternative representations that cover entire scenes, would be valuable. Second, large-scale training across diverse datasets, including both indoor and outdoor scenes and a wide range of object categories, could lead to a generalizable large reconstruction model. Finally, while we currently model static 3D scenes, extending the framework to 4D Gaussians to capture scene dynamics would enable broader applications in dynamic scene understanding and synthesis.

\end{document}